\documentclass[lettersize,journal]{IEEEtran}
\usepackage{amsmath,amsfonts}
\usepackage{algorithmic}
\usepackage{array}
\usepackage[caption=false,font=normalsize,labelfont=sf,textfont=sf]{subfig}
\usepackage{textcomp}
\usepackage{stfloats}
\usepackage{url}
\usepackage{verbatim}
\usepackage{graphicx}
\usepackage{cite}
\usepackage{mathtools}
\usepackage{tabularx}
\usepackage{multirow}
\usepackage{booktabs}
\usepackage[version=4]{mhchem} 
\usepackage{siunitx}
\usepackage[ruled, lined, linesnumbered, commentsnumbered, longend]{algorithm2e}
\usepackage{setspace}
\usepackage{xcolor}

\SetCommentSty{mycommfont}

\SetKwInOut{KwIn}{Input}
\SetKwInOut{KwOut}{Output}

\let\oldnl\nl% Store \nl in \oldnl
\newcommand{\nonl}{\renewcommand{\nl}{\let\nl\oldnl}}% Remove line number for one line

\newlength\mylen
\newcommand\myinput[1]{%
  \nonl
  \settowidth\mylen{\KwIn{}}%
  \setlength\hangindent{\mylen}%
  \hspace*{\mylen}#1
  \\ }

\makeatletter
\newcommand{\doublehat}[1]{% 
\begingroup%
  \let\macc@kerna\z@%
  \let\macc@kernb\z@%
  \let\macc@nucleus\@empty%
  \hat{\raisebox{.2ex}{\vphantom{\ensuremath{#1}}}\smash{\hat{#1}}}%
\endgroup%
}
\makeatother

\newcommand{\specialcell}[2][c]{%
  \begin{tabular}[#1]{@{}c@{}}#2\end{tabular}}

\newcommand\sza{0.235}
\newcommand\szb{0.08em}

\DeclareMathOperator*{\argmax}{arg\,max}
\DeclareMathOperator*{\argmin}{arg\,min}

\begin{document}

\title{Compressive Ptychography using Deep Image and Generative Priors}

\author{Semih Barutcu, Do\u ga G\" ursoy, Aggelos K. Katsaggelos
\thanks{
Semih Barutcu (\textit{Corresponding Author}) is with Department of Electrical and Computer Engineering, Northwestern University in Evanston, IL 60201 USA (semihbarutcu@u.northwestern.edu).}
\thanks{
Doga Gursoy is with X-ray Science Division, Argonne National Laboratory, Lemont, IL 60439 USA, and also with Department of Electrical Engineering, Northwestern University in Evanston, IL 60201 USA (dgursoy@anl.gov).}
\thanks{
Aggelos K. Katsaggelos is with Department of Electrical and Computer Engineering, Northwestern University in Evanston, IL 60201 USA, and is also with Department of Computer Science Engineering, Northwestern University in Evanston, IL 60201 USA (a-katsaggelos@northwestern.edu)
}}
\maketitle

\begin{abstract}

Ptychography is a well-established coherent diffraction imaging technique that enables non-invasive imaging of samples at a nanometer scale. It has been extensively used in various areas such as the defense industry or materials science. One major limitation of ptychography is the long data acquisition time due to mechanical scanning of the sample; therefore, approaches to reduce the scan points are highly desired. However, reconstructions with less number of scan points lead to imaging artifacts and significant distortions, hindering a quantitative evaluation of the results. To address this bottleneck, we propose a generative model combining deep image priors with deep generative priors. The self-training approach optimizes the deep generative neural network to create a solution for a given dataset. We complement our approach with a prior acquired from a previously trained discriminator network to avoid a possible divergence from the desired output caused by the noise in the measurements. We also suggest using the total variation as a complementary before combat artifacts due to measurement noise. We analyze our approach with numerical experiments through different probe overlap percentages and varying noise levels. We also demonstrate improved reconstruction accuracy compared to the state-of-the-art method and discuss the advantages and disadvantages of our approach. 

\end{abstract}

\section{Introduction}

Ptychography is an important computational imaging tool in materials science that enables imaging of large field-of-views at a sub-\SI{10}{\micro\meter} spatial resolution \cite{gursoy2021lensless}. In ptychography, a focused beam of x-ray is scanned across a sample while recording diffraction images in the far-field plane. The images of the sample are obtained after data collection computationally by solving the problem known as phase retrieval \cite{Fienup:78}. Because ptychography uses computing to form an image instead of a physical lens, it avoids the limitations of lens-based imaging systems because of the low photon efficiency of the lenses at hard x-ray energies or the lens-induced distortions in the acquired images. However, despite its high photon efficiency and imaging resolution, the time to obtain images is still long compared with the conventional x-ray microscopes \cite{de2021fast}. The main reason for its slow operation is the mechanical scanning procedure, which requires raster scanning of a micrometer-sized beam of illumination across the sample. For example, while small samples are routinely imaged with ptychography, 3D imaging or imaging of large samples is still challenging \cite{Barutcu2020, du2021upscaling}. Improving the data acquisition time can enable a range of capabilities that are critical for materials science studies, including capturing dynamic phenomena under \emph{operando} conditions or imaging larger sample volumes. 

Traditionally, the ptychographic inverse problem is solved by iterative methods such as the Difference Map (DM) \cite{Thibault_Dierolf_Menzel_Bunk_David_Pfeiffer_2008a, thibault2009probe, du2020three} and extended Ptychographical Iterative Engine (ePIE) \cite{Maiden_Johnson_Li_2017}, and reconstruction quality is increased by more advanced algorithms \cite{bouman}. Initial efforts to improve the scanning time have been concentrated on the development of high-precision scanners that can reliably work at high scan velocities \cite{Pelz:14, Deng:15a, Clark:14}. Alternatively, one can scan fewer data points and use compressive sensing \cite{Candes:06, Donoho:06} to complement the sparsely acquired diffraction images.
%\textcolor{red}{Doga: Cite traditional compressive sensing literature applied to ptychography here and explain the issues.} 
\cite{Moravec2007_pr_theo1, Ohlsson2012_pr_theo2, Newton2012_pr_theo3} provide theoretical limits for lowering the amount of data collected by manipulation of Fourier coefficients, usage of lifting techniques, or employing iterative reweighing for phase retrieval problems. \cite{daSilva2015_decomp} utilizes Gabor's decomposition into elementary signals and \cite{Stevens2018_subsamp_ptycho} lowers the sampling by random removal of detector pixels and diffraction data as examples of the application of compressed sensing to ptychography. However, none of these approaches can produce decent solutions at extreme compression levels, like when the overlapping requirement is completely ignored or when the noise is significantly high due to a dim source or a low exposure time.

To overcome the weaknesses of the traditional ptychographic phase retrieval, deep learning has emerged as a promising alternative or a complementary approach \cite{Sinha2017, Cherukara2018, Xue2019, zhou2020diffraction, Barbas2, Barbas3, wohl}. Self-optimization approaches such as the deep image prior (DIP) \cite{Ulyanov_Vedaldi_Lempitsky_2020} have been effective for phase retrieval \cite{Shamshad_Ahmed_2018, VanRullen_Reddy_2019a, Shamshad_Ahmed_2021, Barbas1}. Its effectiveness is shown also for ptychographic inverse problem in \cite{Yang2020}. In addition, supervised approaches such as the deep generative priors (DGPs) have also been applied for reducing imaging artifacts associated with undersampling \cite{Aslan_Liu_Nikitin_Bicer_Leyffer_Gursoy_2021, Pan_Zhan_Dai_Lin_Loy_Luo_2021}, as well as to overcome issues related to partial-coherence of the source \cite{condmat6040036}. Successful reconstructions using generative networks are demonstrated in \cite{Nguyen2018-cGAN} for Fourier ptychography and \cite{osti_1599580_ptychnet} for x-ray ptychography. 
%\textcolor{blue}{Doga: Are we covering all the important papers here? Semih: I tried to limit these to generative networks to avoid diverging from the main idea. I added a couple more for both DIP and DGP, is that better?}. 
However, both approaches have limitations. On the one hand, the convergence behavior of a DIP approach is unpredictable because it is heavily dependent on the initial state of the network. Also, the improvement expected from an unsupervised method is limited compared to a supervised method. On the other hand, the success of supervised methods such as DGPs is limited by the representation capability of the pre-trained network. Especially when working with sub-optimal training data, DGPs can yield undesired solutions. 

% Because the features of the image can be expressed with much less parameters in which a pre-trained deep network can capture, compression levels beyond the capability of traditional compressive sensing approaches might be possible.

In this paper, to address the weaknesses of both supervised and unsupervised methods for iterative reconstruction for x-ray ptychography, we suggest complementing DIPs with DGPs. We suggest that DGPs can be used as a good initialization of DIPs. We show that using a pre-trained generator network on a specific dataset to limit the generator to the target domain would significantly improve the reconstruction quality. The process is done in two major steps: first, a random vector is optimized to get the best reconstruction in the limited representation capability of the network learned from the pre-training. Then, the network weights are optimized for the target reconstruction by expanding the representation capability towards the required result. In addition to combining DIPs and DGPs, we suggest using iterative updates of the weights with a progressive approach for achieving linear convergence. Furthermore, we suggest the addition of priors such as total variation prior for piecewise-constant smoothness and discriminative prior to penalize divergence from the target domain constrained by the pre-trained network, which is particularly useful for avoiding artifacts in the reconstructions from the noisy data. We validate our method with numerical tests simulating a compressive x-ray ptychography experiment. Since the DGP is trained to generate samples in the specified and trained domain, it enables us to complete the missing data in the low overlap x-ray ptychography. Moreover, we demonstrate the effect of regularization on the reconstructed images to evaluate the artifacts. Finally, we argue that good quality reconstructions may be possible with data acquired with no overlap during scanning, enabling a fully flexible data acquisition with no hard restriction.

\section{PtychoGAN}

In this section, we present the ptychographic phase retrieval problem, describe the components that we suggest using in our model, and describe our approach for solving the problem.

% \subsection{Phase Retrieval}

% In ptychography, we record the intensity measurements in the far-field plane. The phase retrieval problem is to recover the phase $\Phi(k)$ of the complex signal $A(k)exp(j\Phi(k))$ in the detector plane using only the magnitude information $A(k)$. Phase retrieval algorithms aim at iteratively solving for the phase of the signal by imposing some prior knowledge. While some algorithms assumes phase-only objects such as Gerchberg-Saxton Error Reduction algorithm \cite{gerchberg1972practical}, some algorithms assumes a finite spatial extent object \cite{fienup1978reconstruction, fienup1982phase, fienup1987reconstruction}. On the other hand, ptychography is created to avoid these problems, where the required constraints come from the overlapping scanning pattern on the object

\subsection{The Forward Problem}

\begin{figure}[t]
    \centering
    \includegraphics[width=0.7\columnwidth]{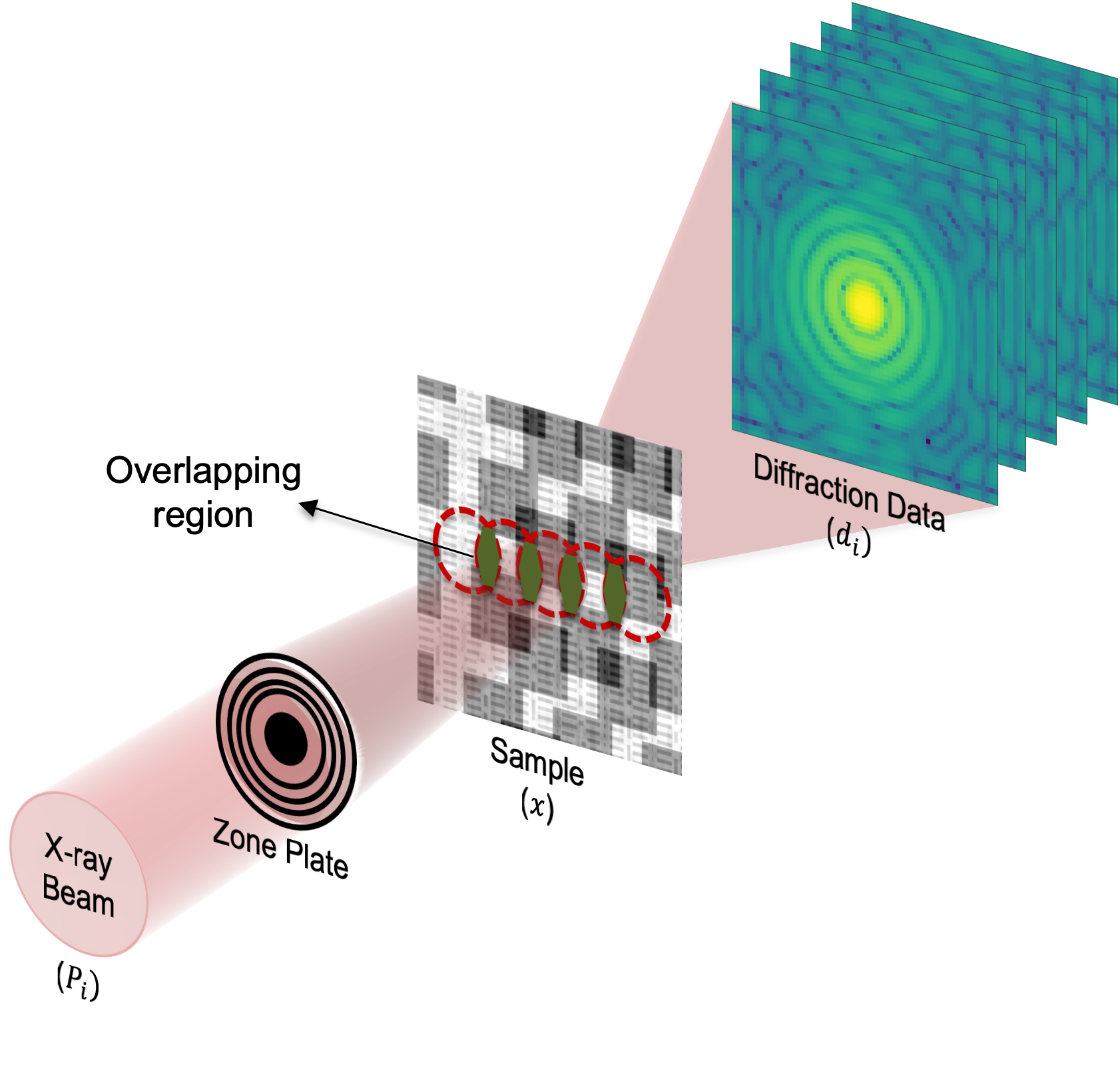}
    \vspace{-15pt}
    \caption{X-ray ptychography experimental setup and overview of the scanning process. A focused beam of x-rays is scanned across the sample by maintaining an overlap in the illumination of consecutive scan points (shown in red). The intensity of the diffracted wave at each scan point is then recorded with a pixel-area detector.}
    \label{fig:ptycho}
\end{figure}

In ptychography, we record the intensity measurements in the far-field plane while scanning the object with a focused probe (Fig.~\ref{fig:ptycho}). Assume that $x \in \mathbb{C}^{N\times N}$ is the unknown 2D complex object of size $N\times N$, and the values of $x$ to depend on the material and the energy of the x-ray source to be scanned. The acquisition process can be described in multiple steps. The object first interacts with the complex illumination function $P \in \mathbb{C}^{M\times M}$ where $M\leq N$. The resulting wave $\psi \in \mathbb{C}^{M\times M}$ in the sample plane is written as,
\begin{equation}\label{Eq:forward1}
     \psi_i = P_i \odot x
\end{equation} 
where we use $i$ to represent the position of the probe in the sample plane, and $\odot$ denotes the element-wise multiplication. Then the wave goes through far-field diffraction represented by the Fourier transform operator $F \in \mathbb{C}^{M\times M}$. The intensity is recorded at the detector, which is represented by the $|.|^2$ operator. 
\begin{equation}
\label{Eq:forward2}
     d_i = |F \psi_i|^2 + \eta_i
\end{equation} 
where $d_i \in \mathbb{R}^{M\times M}$ is the intensity image recorded in the detector, and $\eta_i$ is the additive noise at the scan position $i$.  

\subsection{Ptychographic Inverse Problem}

The phase retrieval problem is to recover the phase $\Phi(k)$ of the complex signal $A(k)exp(j\Phi(k))$ in the detector plane using only the magnitude information $A(k)$. Phase retrieval algorithms aim at iteratively solving for the phase of the signal by imposing some prior knowledge. While some algorithms assume phase-only objects, such as Gerchberg-Saxton Error Reduction algorithm \cite{gerchberg1972practical}, some algorithms assume a finite spatial extent object \cite{fienup1978reconstruction, fienup1982phase, fienup1987reconstruction}. On the other hand, ptychography is created to avoid these problems, where the required constraints come from the overlapping scanning pattern on the object. We can express the phase retrieval minimization problem as:
\begin{equation}
\label{Eq:L1}
      \hat{x} = \argmin_{x} \sum_i \| d_i - |F(P_i \odot x)|^2\|_1.
\end{equation} 
where $\hat{x}$ is the solution. The redundant information in the overlapping regions is utilized to recover the phase of the object. In traditional methods such as ePIE \cite{Maiden_Johnson_Li_2017}, the object is updated for each scanning position consecutively, similar to a gradient-descent-based optimization. This allows the algorithm to update the overlapped regions more than once and use the updated information to recover lost data. However, high overlap constraints significantly increase the data acquisition time. 

\subsection{Compressive Ptychography with Deep Priors}

Compressed sensing is a method using simultaneous acquisition and compression to reduce acquisition time and storage space, increase transmission efficiency, or decrease processing times of a signal. The goal of compressed sensing is to recover a signal $t \in \mathbb{C}^{P}$ from measurements $y = Et \in \mathbb{C}^{R}$ where $E \in \mathbb{C}^{R\times P}$ is the forward model and $P>>R$. Prior information is necessary to support the solution for recovering the true signal in compressed sensing. We pose the phase retrieval as a compressed sensing problem and focus on combining two state-of-the-art methods: deep generative priors and deep image priors.

\subsubsection{Deep Generative Priors}
\label{sec: dgp}

Generative Adversarial Networks (GANs) \cite{goodfellow2014generative} are proven to be effective in learning a representation of a group of samples from a training set which results in having the capability of generating a sample example that can be included in the same sample group. That is done by a method called adversarial training. GANs consist of two neural networks, a generator ($G(.)$) which learns to generate an image in the desired domain, and a discriminator ($D(.)$) that trains the generator by scoring the generated samples. The training is done in an adversarial structure where the generator aims at "fooling" the discriminator, and the discriminator aims at distinguishing between acceptable and unacceptable generated samples \cite{creswell2018generative}.

If trained successfully, the generator learns a mapping from a low dimensional latent vector $z \in \mathbb{R}^{k}$ to a higher dimensional object space $G(z) \in \mathbb{R}^{NxN}$ where $NxN >> k$. The representation capability of the generator network depends on the training process and the complexity of the higher dimensional object space. An ideal generator should span an adequate distribution $P_z$ mapped from a latent vector $z$.

In this paper, we suggest using deep generative priors to reconstruct the object scanned in the x-ray ptychography. Replacing the phase of the unknown object $x$ in Eq.~\ref{Eq:L1} with the output of the generator $G(.)$, and setting the magnitude of the object to 1, the minimization problem becomes:
\begin{equation}\label{Eq:L1_dgp}
      \hat{z} = \argmin_{z} \sum_i \| d_i - |F(P_i \odot e^{jG(z)})|^2\|_1,
\end{equation} 
where,
\begin{equation}\label{Eq:L1_dgp_sol}
      \hat{x} = e^{jG(\hat{z})}.
\end{equation}
Therefore, the phase retrieval problem becomes solving for the latent vector $z$ that minimizes Eq.~\ref{Eq:L1_dgp}. To minimize the network's approximation error, the generator should have a significant representation capability, which often requires training on a large dataset of images in the same subset of images with the scanned object. However, this is usually not practical due to the limited availability of open-access datasets for training.

\subsubsection{Deep Image Priors}

Deep networks trained on a limited number of samples will have fundamentally limited effectiveness defined by the training data used. Even if a generator network is trained with a considerably large dataset, the representation capability is still not ideal, and the solution is likely beyond the representation capability of the network. A path forward to combat with the limited representation capability of pre-trained networks is to use untrained networks, i.e., deep image priors. With DIPs, the network architecture itself constrains the solution and allows the reconstruction to fit the measurement data. They do so by leveraging the superior low-dimensional image representation capability of deep generative networks. We propose a method to use deep image priors as a fine-tuning method to get the required reconstructions after the initial reconstruction acquired by the deep generative prior. Based on this approach, Eq.~\ref{Eq:L1_dgp} can be modified as:
\vspace{-5pt}
\begin{equation}\label{Eq:L1_dip}
      \hat{z}, \hat{\theta} = \argmin_{z, \theta} \sum_i \| d_i - |F(P_i \odot e^{jG_{\theta}(z)}|^2\|_1,
\end{equation} 
where,
\vspace{-5pt}
\begin{equation}\label{Eq:L1_dip_sol}
      \hat{x} = e^{jG_{\hat{\theta}}(\hat{z})},
\end{equation}
and $\theta$ represents the weights of the network. Here the network structure of the generator can be used as the prior, and by optimizing the parameters of the network, we can achieve an improved reconstruction of the object.

\subsection{Progressive Adjustment Optimization}

Optimization of the weights of a large network for a specific problem is not straightforward. The function is bound to many variables is non-convex, and minimization is challenging. To avoid this problem, we suggest using progressive optimization in which we update the parameters of the generator network starting from the shallowest layer. Then, the number of updated layers is increased progressively until the whole network weights are completely updated.

The progressive adjustment allows the network to focus on different layers in a sequential manner and to improve the reconstruction quality by minimizing the possibility of being stuck at a local minimum. We denote the weight of the shallowest layer of a network by $\theta_0$, and the total number of layers by $L$. The process of updating weights can be shown by Algorithm~\ref{alg:alg1}.

\begin{algorithm}
\setstretch{1.1}
\caption{Progressive Optimization}%\label{alg:alg0}
\label{alg:alg1}
% \begin{algorithmic}
% \STATE 
\For {$l$ in $L$}{
    $\hat{z}, {\hat{\theta_0}, ..., \hat{\theta_l}}$
    $= \argmin\limits_{z, {\theta_0, ..., \theta_l}} \sum_i \| d_i - |F(P_i \odot e^{jG_{\theta}(z)}|^2\|_1$
}
\KwRet $e^{jG_{\hat{\theta_0}, ...,\hat{\theta_L}}(\hat{z})}$
% \STATE return  
% \end{algorithmic}
\end{algorithm}

% \vspace{-20pt}
\subsection{Regularization}

Prior information about the target object is crucial for an accurate solution to the phase retrieval problem. Regularization applied to the inverse problem can significantly increase the reconstruction quality depending on the specific problem or enable reconstructions with limited or noisy data. Total variation regularization and discriminative regularization are the regularization types that we apply as part of our method.

\subsubsection{Total Variation Regularization}

For many objects in x-ray ptychography piece-wise continuity is a common feature to have as the objects scanned usually consist of artificial structures. The information available in this form can be a valuable prior to reduce artifacts in the reconstruction and to force the regularizer to a better solution. In addition, total variation minimization can leverage the solution for noisy data significantly by eliminating the sparse artifacts. For the x-ray ptychography minimization we suggest, the total variation distance can be defined as:
\begin{equation}
\label{Eq:tv_reg}
      TV(z, \theta) = \| \nabla (G_{\theta}(z)) \|_1
\end{equation} 
where $\nabla x$ designates the gradient operation on the object. Adding the regularization term to our minimization function in Eq.~\ref{Eq:L1_dip} with the regularization parameter $\lambda_1$ controlling the trade-off between the data fidelity and regularization parts of the function, the minimization can be expressed as:
\begin{multline}
\label{Eq:reg1_added}
      \hat{z}, \hat{\theta} = \argmin_{z, \theta} \sum_i \| d_i - |F(P_i \odot e^{jG_{\theta}(z)}|^2\|_1 \\ + \lambda_1 \| \nabla (G_{\theta}(z)) \|_1.
\end{multline}

\subsubsection{Discriminative Regularization}

As described in Section~\ref{sec: dgp}, DGPs are strong priors to drive the reconstruction closer to the true solution. However, the addition of DIPs can drive the solution away from the true solution due to missing information and the added noise in measurements. To avoid divergence from the true solution and getting trapped in an undesired local minimum, we employ a discriminative regularization in our DGP approach.

The network used for the optimization is the trained generator network of a GAN structure, while the discriminator is discarded. However, the discriminator is valuable as it is trained to distinguish if the reconstruction from the generator belongs to the same domain as the objects from the training set. Thus, the loss from the discriminator output can be used to create a regularizer punishing the divergence from the required domain. $D(.)$ describing the discriminator, the discriminator loss (DL) can be written as:
\begin{equation}
\label{Eq:disc_loss}
      DL(z, \theta) = log (1 - D((G_{\theta}(z)).
\end{equation} 
The discriminator loss can be added to the minimization function in Eq.~\ref{Eq:reg1_added} with the regularization parameter $\lambda_2$ as:
\begin{multline}
\label{Eq:reg2_added}
      \hat{z}, \hat{\theta} = \argmin_{z, \theta} \sum_i \| d_i - |F(P_i \odot e^{jG_{\theta}(z)}|^2\|_1 \\ + \lambda_1 \| \nabla (G_{\theta}(z)) \|_1 + \lambda_2 log (1 - D((G_{\theta}(z)).
\end{multline}

\subsection{Bayesian Approach}

Photon detection by a detector can be described as a Poisson process. Therefore, the probability of measuring the diffraction image $d_i$, can be expressed by the likelihood function. With the assumption that the measurements are independent of each other, instead of the minimization in Eq.~\ref{Eq:L1}, we can maximize the probability of measuring all data points.

\begin{eqnarray}
\label{Eq:d_given_x}
    \hat{x} &=& \argmax_{x} p(d|x) \\
    &=& \argmax_{x} \prod_{i} \frac{e^{|F(P_i \odot x)|^2} |F(P_i \odot x|^{2d_i}}{d_i!}
\end{eqnarray} 
Here, we assume that we have a prior knowledge on the object to be reconstructed. So, instead, we can maximize the posteriori probability $p(x|d)$ which can be calculated by the Bayes' theorem:
\begin{multline}
\label{Eq:bayes}
    \hat{x} = \argmax_{x} p(x|d) = \argmax_{x} \frac{p(d|x)p(x)}{p(d)}
    \\ \hspace{-28pt} = \argmax_{x} {p(d|x)p(x)} 
    \\ = \argmin_{x} (-log(p(d|x))-log(p(x)))
\end{multline} 
where,
\begin{equation}
\label{Eq:log_d_given_x}
    -log(p(d|x)) = \sum_{i} (F(P_i \odot x)|^2 - 2d_i log(|F(P_i \odot x)|).
\end{equation}
As we have previously discussed, we would like to apply piece-wise continuity to the reconstruction to suppress imaging noise. So, we select the object in the Gibbs form, which allows us to apply total variation regularization.
\begin{eqnarray}
\label{Eq:gibbs}
    p(x) &=& e^{\lambda_1 \| \nabla (x) \|_1} \\
    -log(p(x)) &=& \lambda_1 \| \nabla (x) \|_1.
\end{eqnarray}
Combining these information, maximum a posteriori probability (MAP) estimate for the x-ray ptychography problem can be written as the minimization:
\begin{multline}
\label{Eq:map}
    \hat{x} = \argmin_{x} \sum_{i} (|F(P_i \odot x)|^2 - 2d_i log|F(P_i \odot x)|) 
    \\ + \lambda_1 \| \nabla (x) \|_1
\end{multline}
which can be used to update our minimization equation Eq.~\ref{Eq:reg2_added}:
\begin{eqnarray}
\label{Eq:converted}
    \hat{z}, \hat{\theta} &=& \argmin_{z, \theta} \sum_i \| |F(P_i \odot e^{jG_{\theta}(z)}|^2 \nonumber
    \\ && - 2d_i log|F(P_i \odot e^{jG_{\theta}(z)}|\|_1  \nonumber
    \\ && + \lambda_1 \| \nabla (G_{\theta}(z)) \|_1 \nonumber
    \\ && + \lambda_2 log (1 - D((G_{\theta}(z)).
\end{eqnarray}
\vspace{-15pt}
% \subsection{Proposed Solution for Undersampled Ptychography}
\begin{algorithm} [b]
\setstretch{1.00}
\caption{Proposed Solution}
\label{alg:alg2}

% \hspace*{\algorithmicindent} \textbf{Input:} $z \in \mathbb{R}^{k}$, a random vector \\
% % \hspace*{\algorithmicindent}
% \hspace*{\algorithmicindent} \textbf{Output:}  $G(z) \in \mathbb{R}^{NxN}$

\KwIn{$z \in \mathbb{R}^{k}$, a random vector}
\myinput{$G_{\theta}(.) \in \mathbb{R}^{NxN}$, a generator network}
\myinput{$D_{\omega}(.) \in \mathbb{R}^{1}$, a discriminator network}
\myinput{$I_{train}$, a set of images to train the networks}
\myinput{$d_i \in \mathbb{R}^{M\times M}$, intensity of diffractions at pos. \textit{i}}
\KwOut{$\hat{x} \in \mathbb{R}^{NxN}$, reconstructed object}

\nonl \tcp{Pre-train the networks}
\nl Train $G_{\theta}(.)$ and $D_{\omega}(.)$ using $I_{train}$
\nonl \tcp{Find an initial reconstruction $x_{init}$}
\nl $\hat{z} = \argmin\limits_{z} \sum_i \| d_i - |F(P_i \odot e^{jG_{\theta}(z)}|^2\|_1$
\\
$x_{init} = e^{jG_{\theta}(\hat{z})}$
\\
\nonl \tcp{Improve reconstruction quality}
\nonl \tcp{Apply progressive adjustment (Alg.~\ref{alg:alg1})}
\nl $\doublehat{z}, \hat{\theta} = \argmin\limits_{z, \theta} \sum_i \| |F(P_i \odot e^{jG_{\theta}(z)}|^2 $
\\
\nonl$ \hspace{18pt} - 2d_i log|F(P_i \odot e^{jG_{\theta}(\hat{z})}|\|_1 $
\\
\nonl$ \hspace{18pt} + \lambda_1 \| \nabla (G_{\theta}(\hat{z})) \|_1 $
\\
\nonl $ \hspace{18pt} + \lambda_2 log (1 - D_{\omega}((G_{\theta}(\hat{z})). $

\KwRet $\hat{x} = e^{jG_{\hat{\theta}}(\doublehat{z})}$

\end{algorithm}

Combining the steps described in previous sections, we can solve the minimization problem of undersampled ptychography in Eq.~\ref{Eq:forward2} with low or no overlap data. The novel algorithm we propose in this paper is shown in Algorithm~\ref{alg:alg2} and can be summarized as:

\begin{itemize}
\item Train generator-discriminator networks with enough data for the network to learn generating samples in the desired output domain.
\item Find an initial reconstruction using Eq.~\ref{Eq:L1_dgp}.
\item Improve reconstruction quality using Eq.~\ref{Eq:converted} while applying Algorithm~\ref{alg:alg1} to update the network weights.
\end{itemize}

% \textcolor{red}{Write this as an algorithm?}

% \textcolor{red}{Doga: Maybe add this at the end of the F. Bayesian ... section and have an explicit pseudo-algorithm to describe the steps.}\textcolor{blue} {I changed it but I am not sure if that is easier or harder to understand.}

\section{Experiments and Results}

In this section, we describe the numerical experiments to validate the proposed algorithm, show the results, and compare the capabilities of the approach with a well-understood and widely used method.

\subsection{Simulation Setup}

\begin{figure}[!b]
    \centering
    \setlength{\tabcolsep}{\szb}

    \begin{tabular}{ccc}
    \multicolumn{3}{c}{\small {Input Images}}
    \vspace{2pt}
    \\
    \includegraphics[width=\sza\columnwidth]{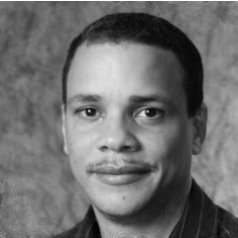} & 
    \includegraphics[width=\sza\columnwidth]{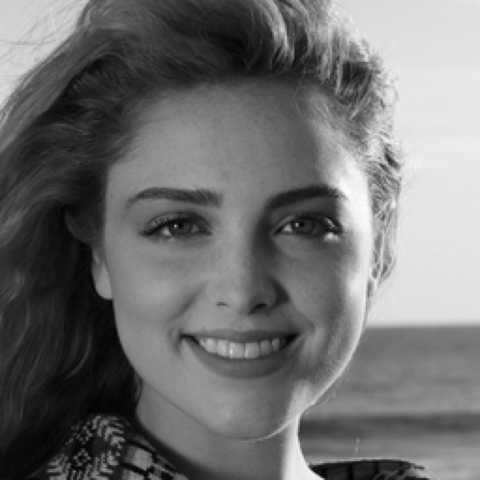} & 
    \includegraphics[width=\sza\columnwidth]{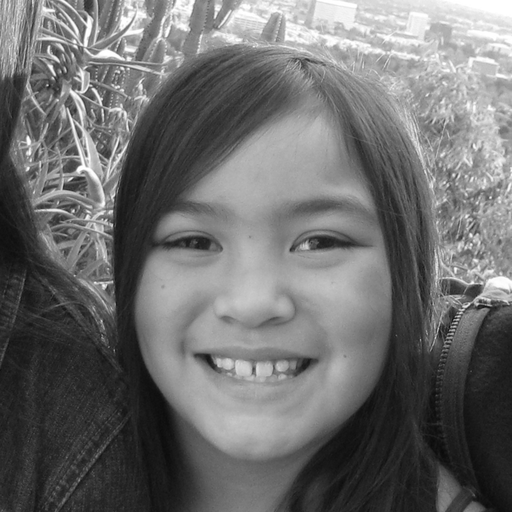}
    \\
    \multicolumn{3}{c}{\small {Diffraction Data Samples and Probe}}
    \vspace{2pt}
    \\
    \includegraphics[width=\sza\columnwidth]{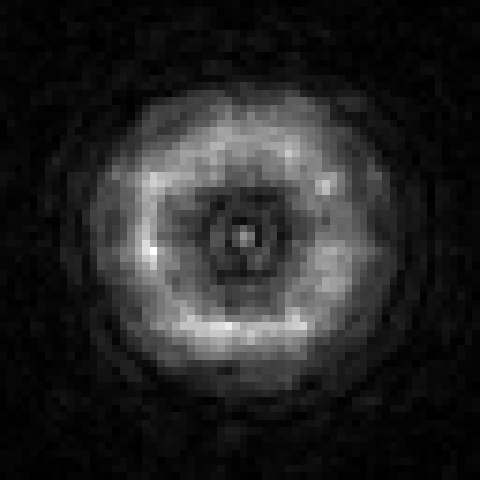} &
    \includegraphics[width=\sza\columnwidth]{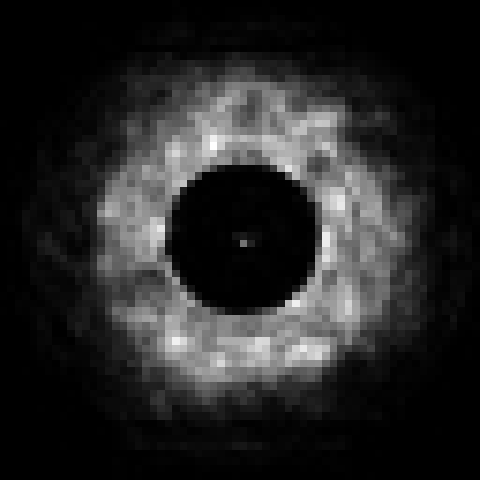} & 
    \includegraphics[width=\sza\columnwidth]{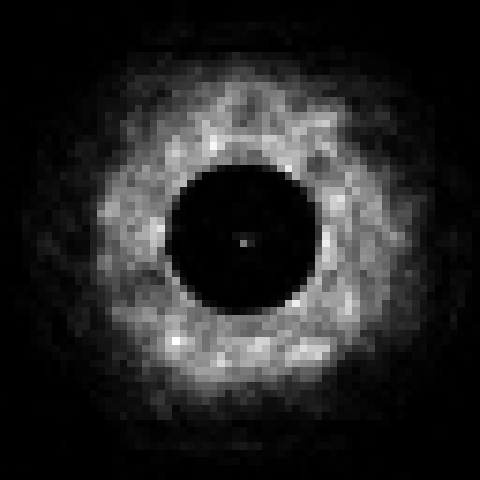}
    \vspace{-4pt}
    \\
    \footnotesize Probe & \footnotesize $\sigma = 0$ & \footnotesize $\sigma = 0.2$
    \\
    \includegraphics[width=\sza\columnwidth]{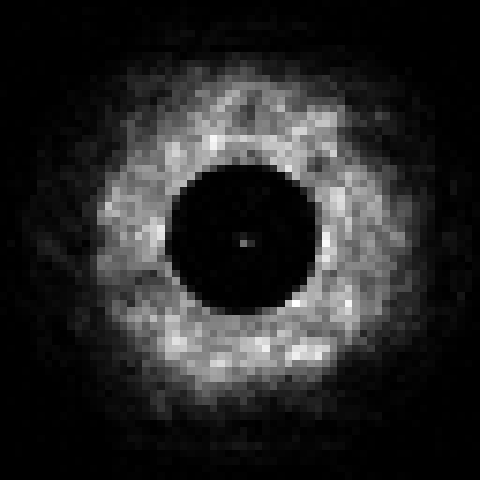} & 
    \includegraphics[width=\sza\columnwidth]{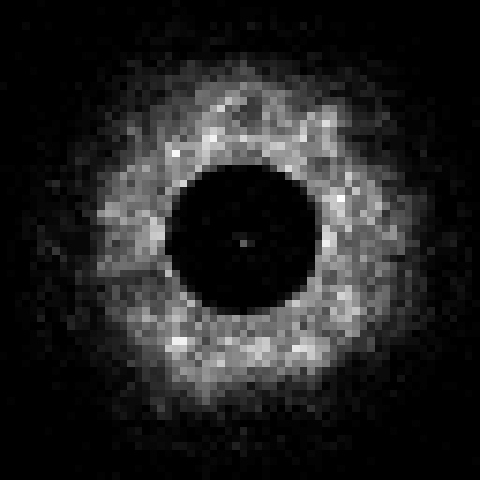} & 
    \includegraphics[width=\sza\columnwidth]{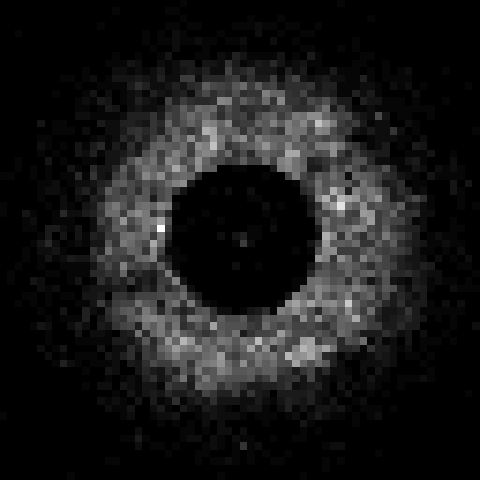}
    \vspace{-4pt}
    \\
    \footnotesize $\sigma = 0.5$ & \footnotesize $\sigma = 2$ & \footnotesize $\sigma = 5$ 
    
\end{tabular}
\caption{Sample input images, phase of the probe used, and sample acquired data in for different noise amounts.}
\label{fig:sample}
\end{figure}
\begin{figure}[hb]
    \centering
    \setlength{\tabcolsep}{\szb}

    \begin{tabular}{cccc}
    {$G_{\theta}(z)$} & {$G_{\theta}(\hat{z})$} & {$G_{\hat{\theta}}(\hat{z})$} & {GT}
    \\
    \includegraphics[width=\sza\columnwidth]{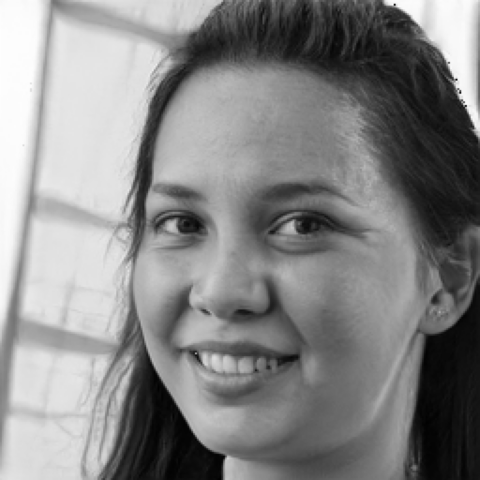} & 
    \includegraphics[width=\sza\columnwidth]{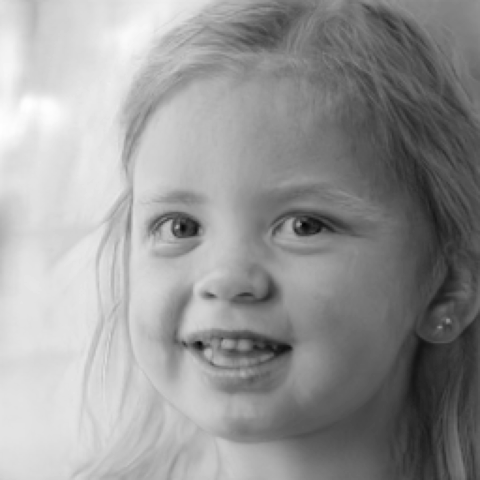} &
    \includegraphics[width=\sza\columnwidth]{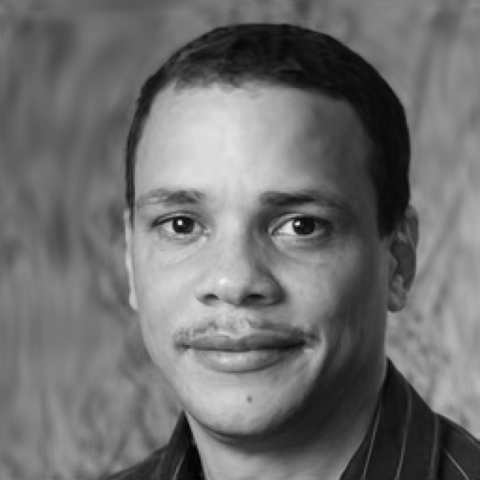} &
    \includegraphics[width=\sza\columnwidth]{figs/gt1.png}
    \\
    \includegraphics[width=\sza\columnwidth]{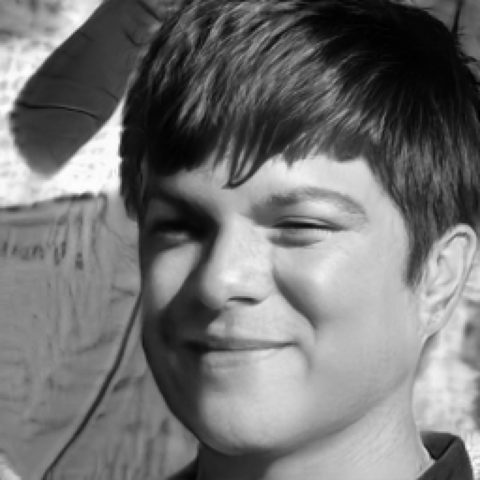} & 
    \includegraphics[width=\sza\columnwidth]{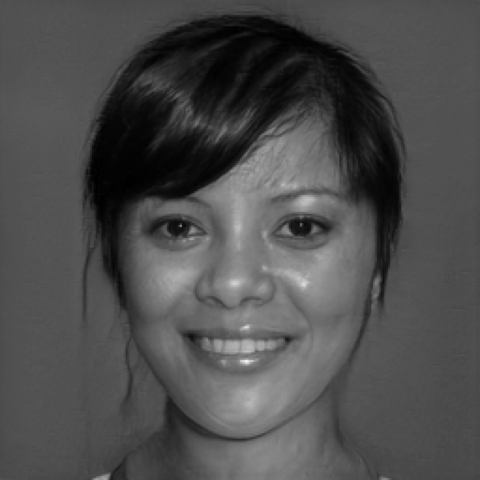} &
    \includegraphics[width=\sza\columnwidth]{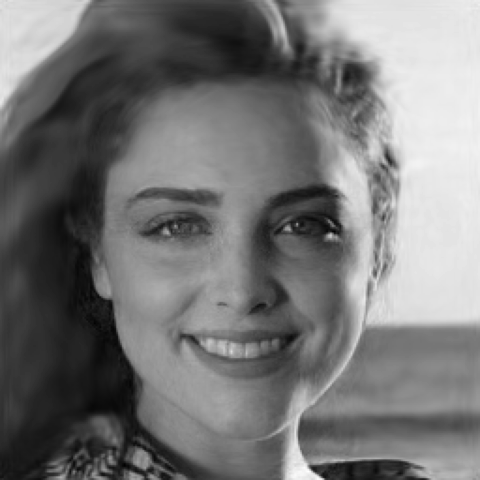} &
    \includegraphics[width=\sza\columnwidth]{figs/gt2.png}
    \\
    \includegraphics[width=\sza\columnwidth]{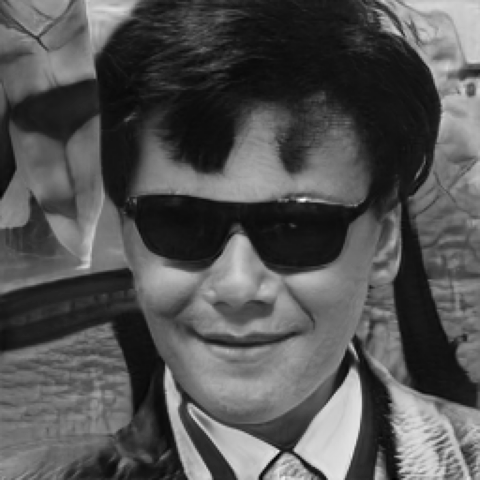} & 
    \includegraphics[width=\sza\columnwidth]{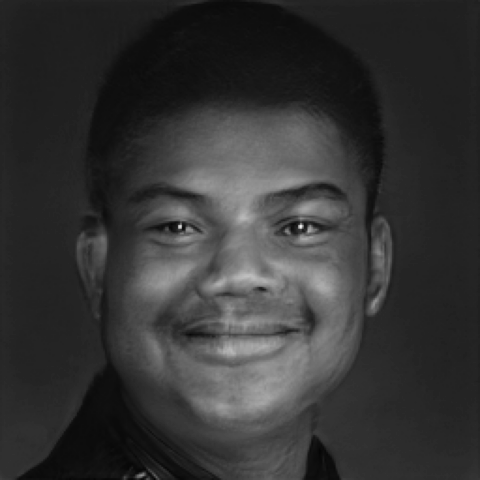} &
    \includegraphics[width=\sza\columnwidth]{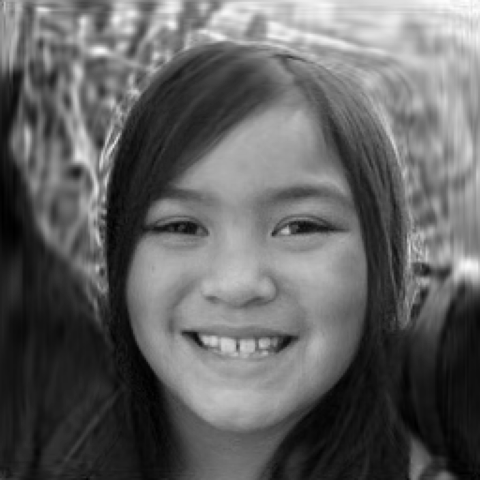} &
    \includegraphics[width=\sza\columnwidth]{figs/gt3.png}

\end{tabular}

\caption{Initial reconstruction without any adjustments, reconstructions after adjustment of latent vector and ground truth (GT) images for 50\% overlap and noise-free cases.}
 \label{fig:rec1_nf}
\end{figure}

\begin{figure*}[!ht]
    \centering
    \setlength{\tabcolsep}{\szb}

    \begin{tabular}{cccccccc}
    {$G_{\theta}(z)$} & {$G_{\theta}(\hat{z})$} & {$G_{\hat{\theta}_{0}}(\hat{z})$} &  {$G_{\hat{\theta}_{0..2}}(\hat{z})$} & {$G_{\hat{\theta}_{0..4}}(\hat{z})$} & {$G_{\hat{\theta}_{0..N}}(\hat{z})$} & {$G_{\hat{\theta}_{no-prog}} (\hat{z})$} & {GT} 
    \vspace{2pt}
    \\
    \includegraphics[width=\sza\columnwidth]{figs/inits/im1_init.png} & 
    \includegraphics[width=\sza\columnwidth]{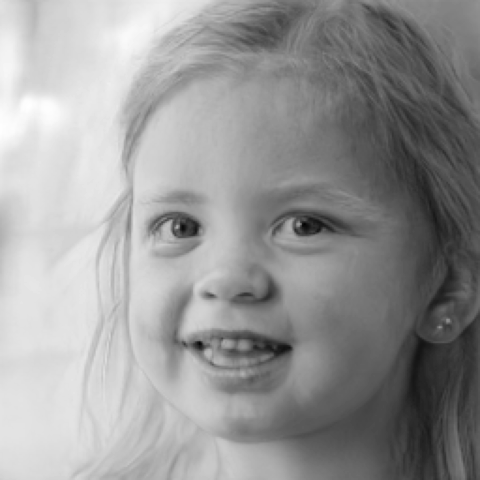} &
    \includegraphics[width=\sza\columnwidth]{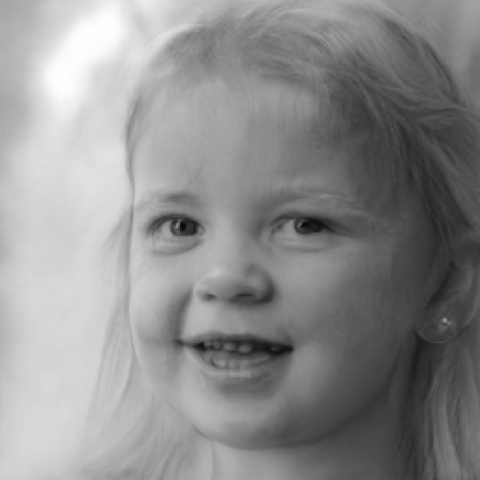} &        \includegraphics[width=\sza\columnwidth]{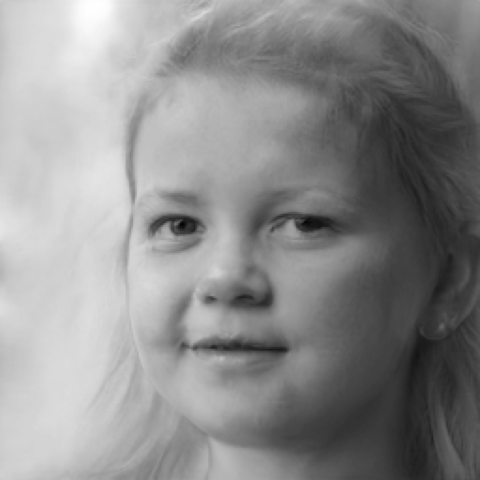} &   
    \includegraphics[width=\sza\columnwidth]{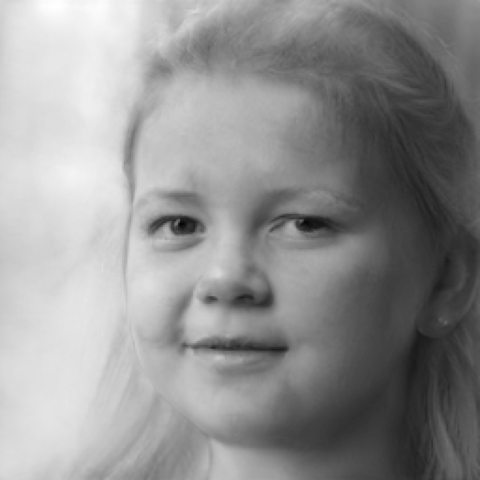} &   
    \includegraphics[width=\sza\columnwidth]{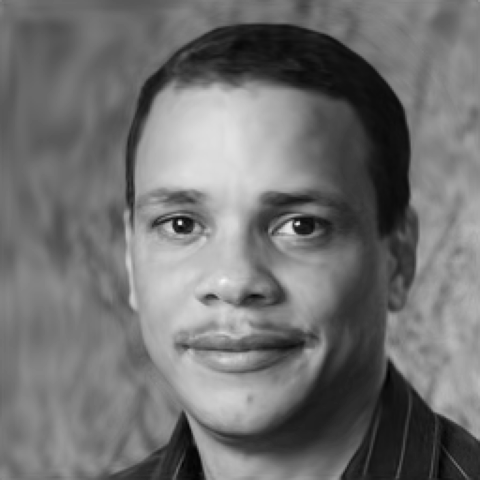} &   
    \includegraphics[width=\sza\columnwidth]{figs/inits/im1_recon.png} &   
    \includegraphics[width=\sza\columnwidth]{figs/gt1.png}
    \\
    \includegraphics[width=\sza\columnwidth]{figs/inits/im1_init.png} & 
    \includegraphics[width=\sza\columnwidth]{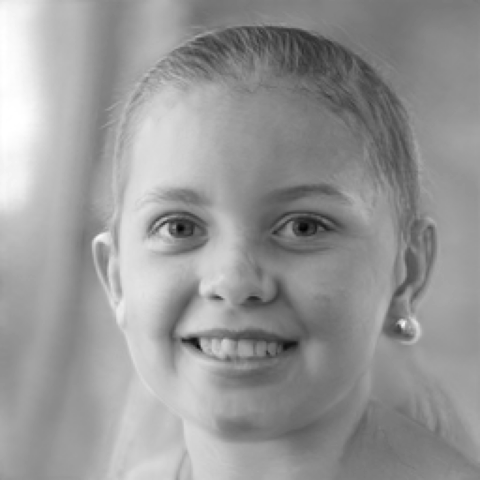} &
    \includegraphics[width=\sza\columnwidth]{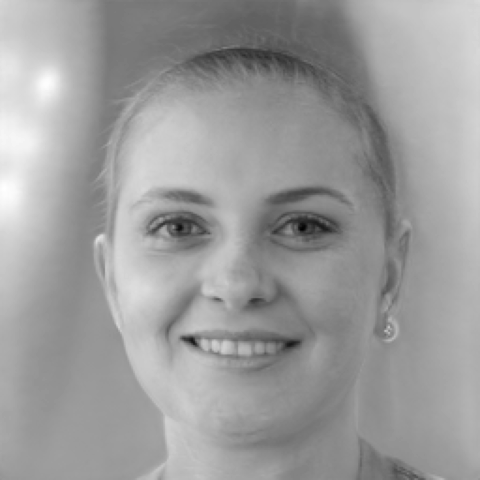} &        \includegraphics[width=\sza\columnwidth]{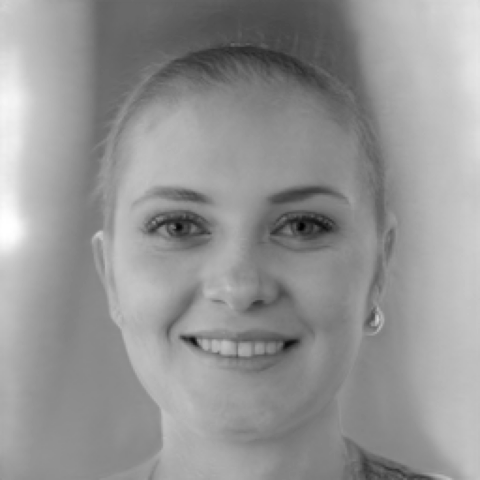} &   
    \includegraphics[width=\sza\columnwidth]{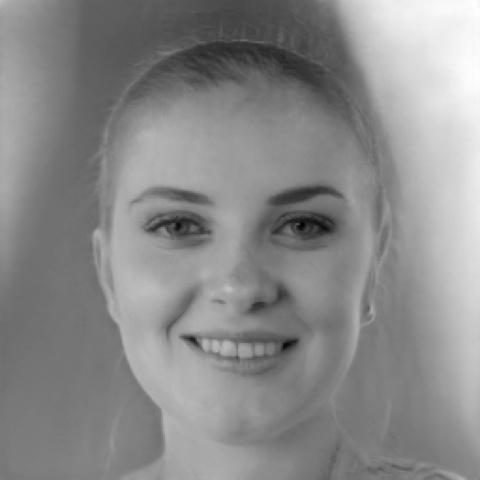} &   
    \includegraphics[width=\sza\columnwidth]{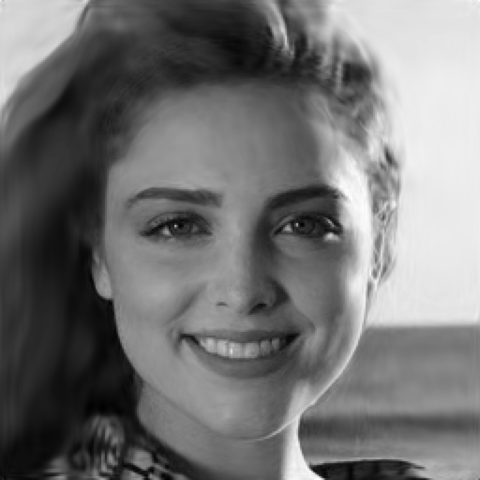} &   
    \includegraphics[width=\sza\columnwidth]{figs/inits/im2_recon.png} &  
    \includegraphics[width=\sza\columnwidth]{figs/gt2.png}

\end{tabular}
\caption{Initial reconstruction without any adjustments, reconstructions after adjustment of latent vector, and reconstructions after adjusting the parameters of the network with progressive adjustment at different levels, ePIE reconstruction, and ground truth (GT) images for 50\% overlap and noise-free cases.}
 \label{fig:rec2_prog_nf}
\end{figure*}

To evaluate the effectiveness of the proposed method against the state-of-the-art methods, numerical tests are performed on face images to represent the idea. Instead of designing generator-discriminator network architecture, the networks structures are taken from the StyleGAN2 architecture \cite{stylegan2}, which is specialized in generating different faces from a random vector. A PyTorch implementation of these networks is used for accurate representation \cite{Wang2020}. 

To utilize the deep generative priors, the networks are trained on a large face dataset called Flickr-Faces-HQ (FFHQ) \cite{stylegan}. The resulting generator network has the capacity to generate different realistic faces from a given random vector, and the discriminator network is capable of classifying the generated images as real or fake. Three images from the test dataset are chosen as the images to be reconstructed, which the network has never seen before. To simulate the measurement data, these images are turned into gray-scale images and converted to complex matrices with the size of $256\times 256$ pixels by setting the complex numbers' phase values to the gray-scale objects and the amplitude values to 1. The probe for the simulations is obtained from the reconstruction of a real object using the algorithm proposed in \cite{Nashed_Peterka_Deng_Jacobsen_2017}, and the diameter of the probe is set to be 32 pixels. With the acquired complex object and the probe, the forward model of x-ray  ptychography described by Eqs~\ref{Eq:forward1} and \ref{Eq:forward2} is applied with varying step sizes and with varying Poisson noise levels according to the peak intensities. The ground truth images and the acquired data without noise and the highest noise level are seen in Fig.~\ref{fig:sample}. As it can be observed, apart from they are all face images for getting advantage of DGPs, the input images are significantly different from each other, and the noisy data has some missing information.

\begin{figure*}[!ht]
    \centering
    \setlength{\tabcolsep}{\szb}

    \begin{tabular}{ccccccccc}
    & {$\sigma = 0.2$} & {$\sigma = 0.5$} & {$\sigma = 2$}  & {$\sigma = 5$}
    & {$\sigma = 0.2$} & {$\sigma = 0.5$} & {$\sigma = 2$}  & {$\sigma = 5$}
    \vspace{2pt}
    \\
    \rotatebox{90}{\hspace{0pt}\emph{\specialcell{\hspace{10pt} DL = 0 \\ \hspace{10pt} TV = 0}}}
    \hspace{1pt} &
    \includegraphics[width=\sza\columnwidth]{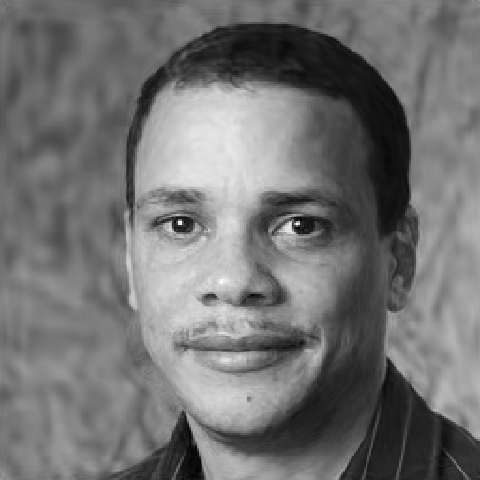} &
    \includegraphics[width=\sza\columnwidth]{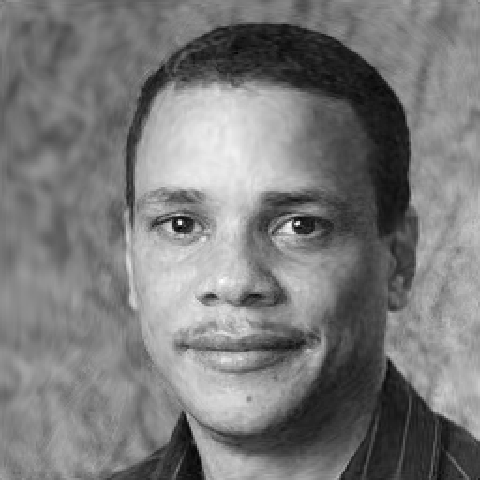} &     \includegraphics[width=\sza\columnwidth]{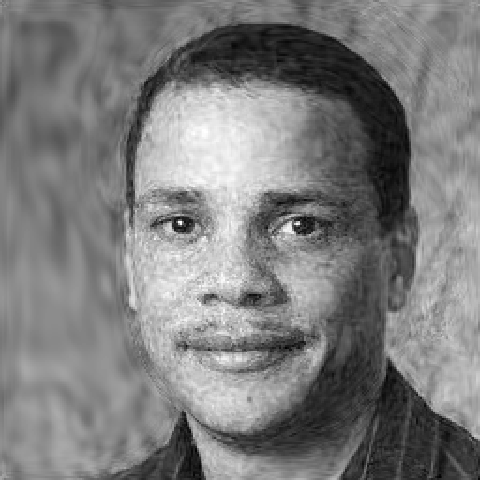} &
    \includegraphics[width=\sza\columnwidth]{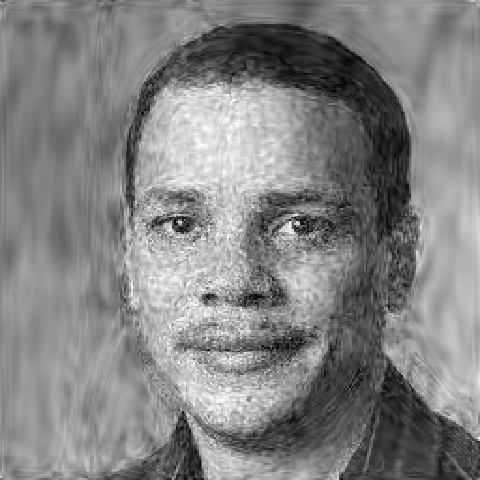} &
    \hspace{0pt}
    \includegraphics[width=\sza\columnwidth]{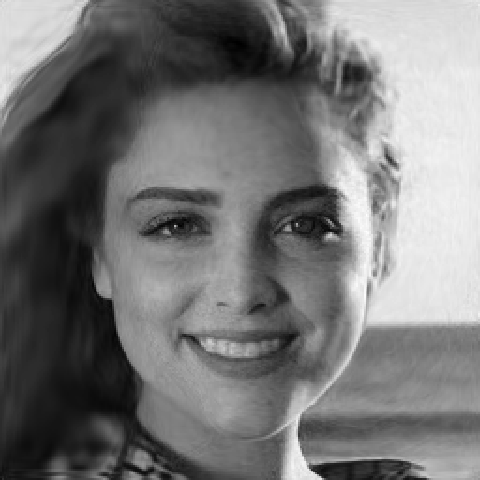} &     \includegraphics[width=\sza\columnwidth]{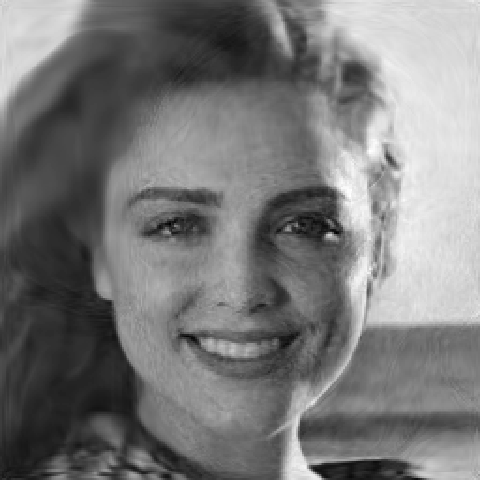} &     \includegraphics[width=\sza\columnwidth]{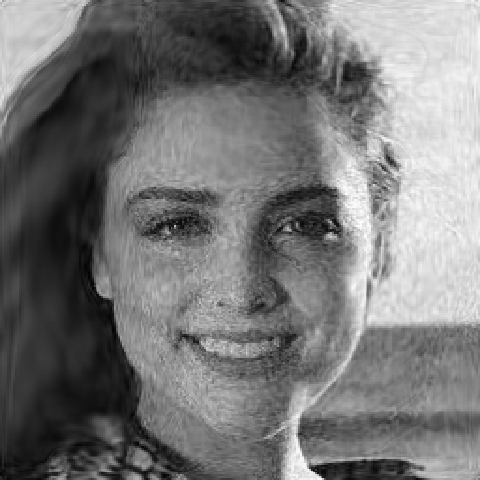} &
    \includegraphics[width=\sza\columnwidth]{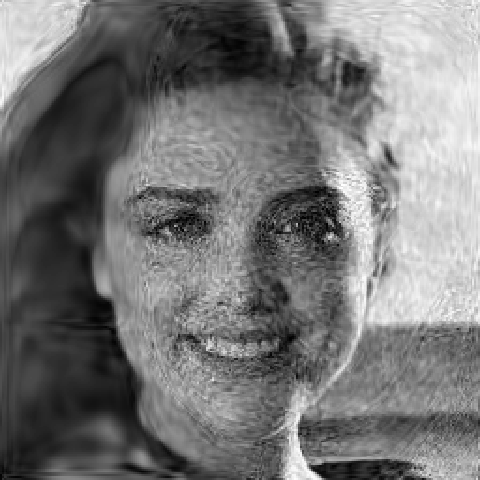}
    \\
    \rotatebox{90}{\hspace{0pt}\emph{\specialcell{\hspace{4pt} DL = 1e-5 \\ \hspace{4pt} TV = 3e-4}}}
    \hspace{1pt} &
    \includegraphics[width=\sza\columnwidth]{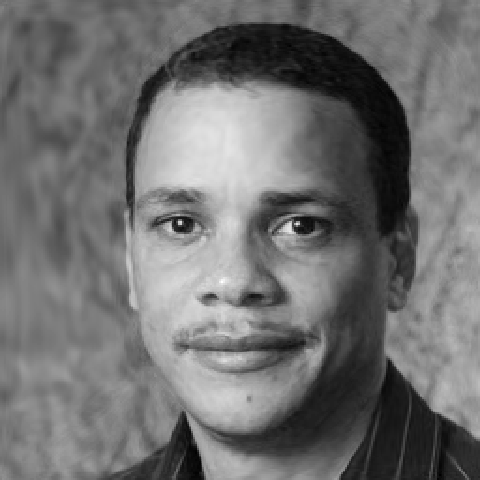} & 
    \includegraphics[width=\sza\columnwidth]{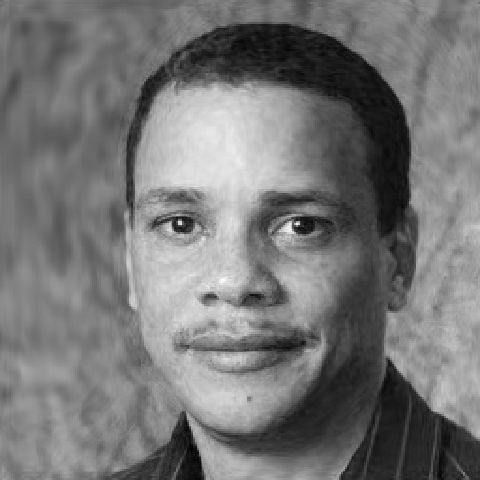} &     \includegraphics[width=\sza\columnwidth]{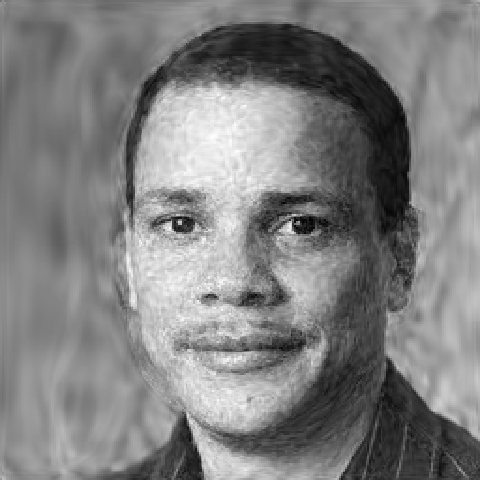} &     \includegraphics[width=\sza\columnwidth]{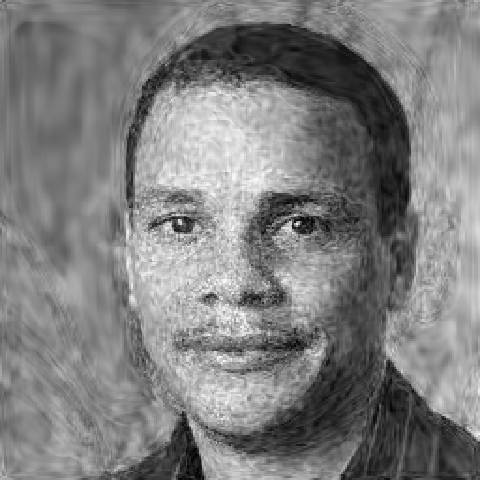} & 
    \hspace{0pt}
    \includegraphics[width=\sza\columnwidth]{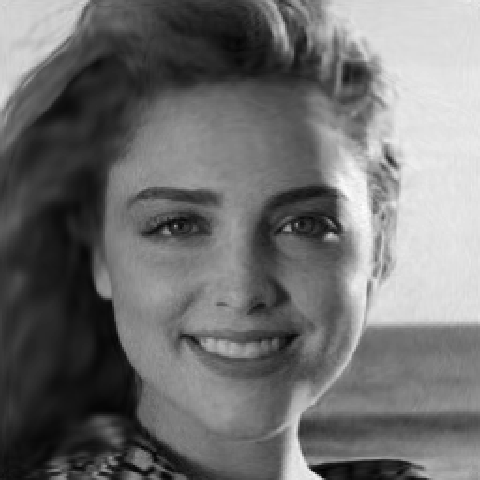} & 
    \includegraphics[width=\sza\columnwidth]{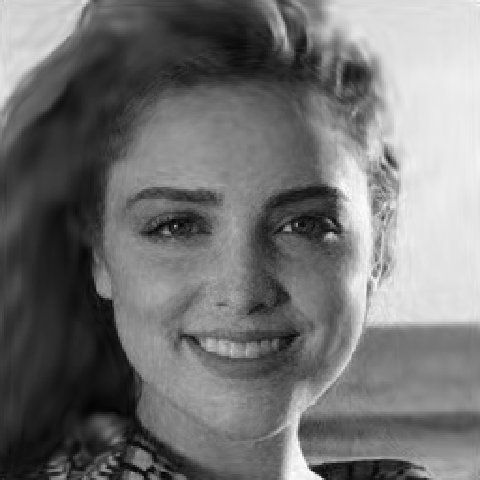} &     \includegraphics[width=\sza\columnwidth]{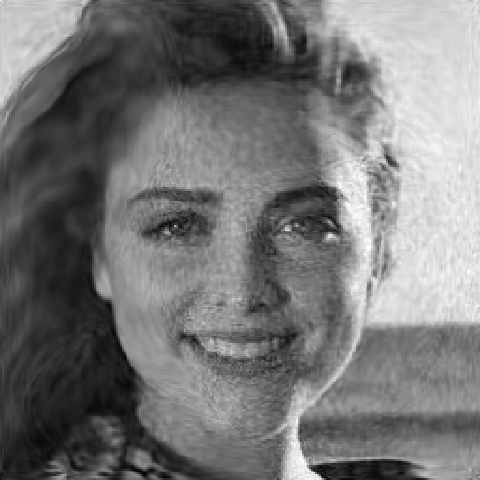} &     \includegraphics[width=\sza\columnwidth]{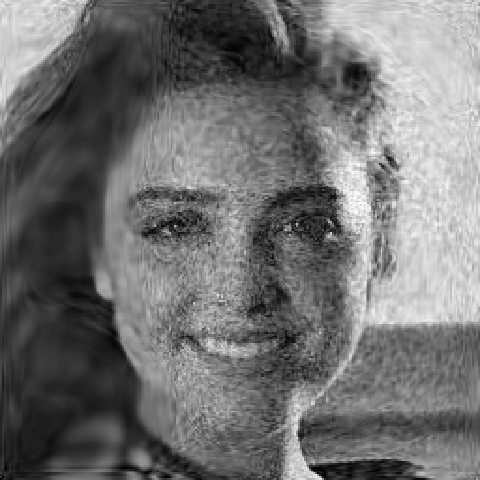}
    \\
    \rotatebox{90}{\hspace{0pt}\emph{\specialcell{\hspace{4pt} DL = 1e-5 \\ \hspace{4pt} TV = 1e-3}}}
    \hspace{1pt} &
    \includegraphics[width=\sza\columnwidth]{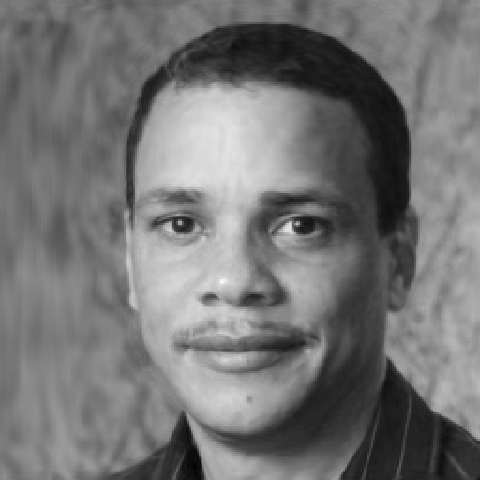} & 
    \includegraphics[width=\sza\columnwidth]{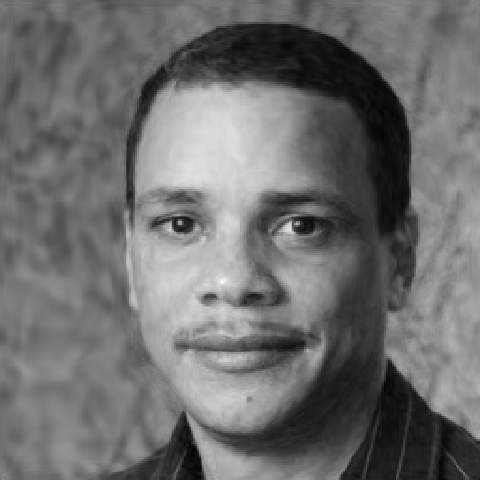} &    \includegraphics[width=\sza\columnwidth]{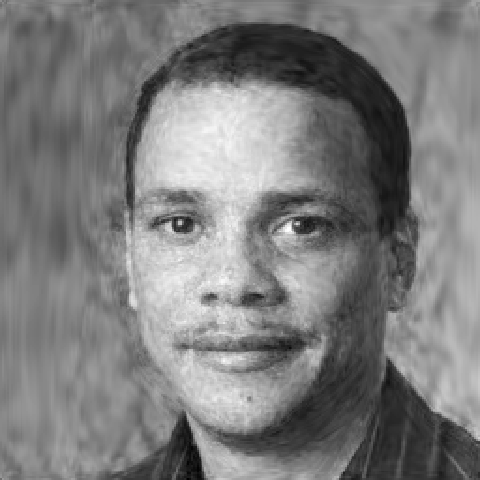} &     \includegraphics[width=\sza\columnwidth]{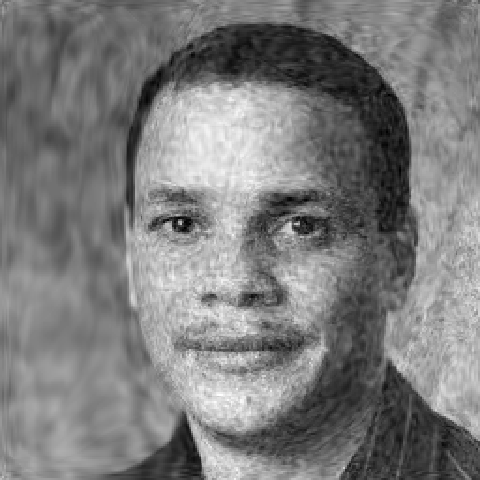} & 
    \hspace{0pt}
    \includegraphics[width=\sza\columnwidth]{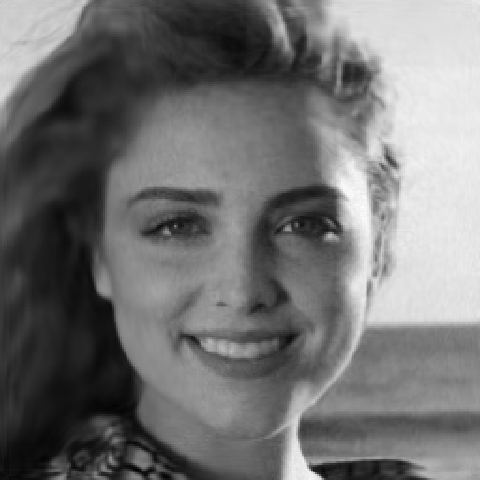} & 
    \includegraphics[width=\sza\columnwidth]{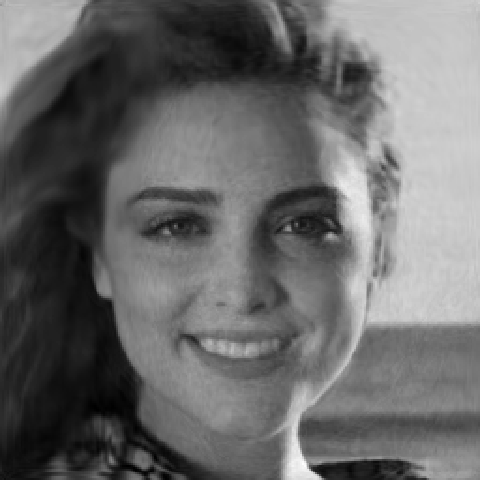} &     \includegraphics[width=\sza\columnwidth]{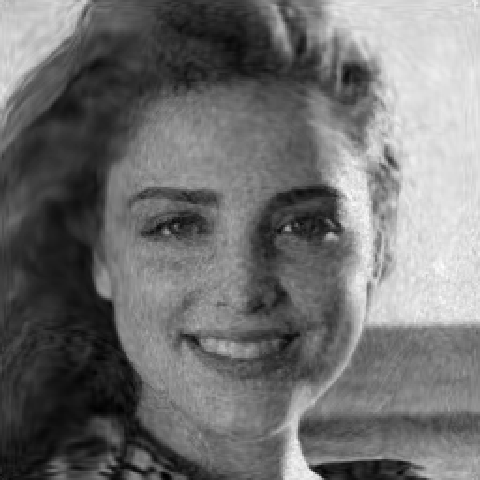} &     \includegraphics[width=\sza\columnwidth]{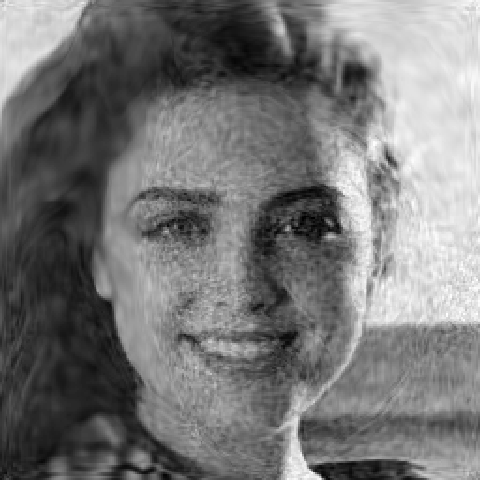}
    \\
    \rotatebox{90}{\hspace{0pt}\emph{\specialcell{\hspace{4pt} DL = 1e-5 \\ \hspace{4pt} TV = 3e-3}}}
    \hspace{1pt} &
    \includegraphics[width=\sza\columnwidth]{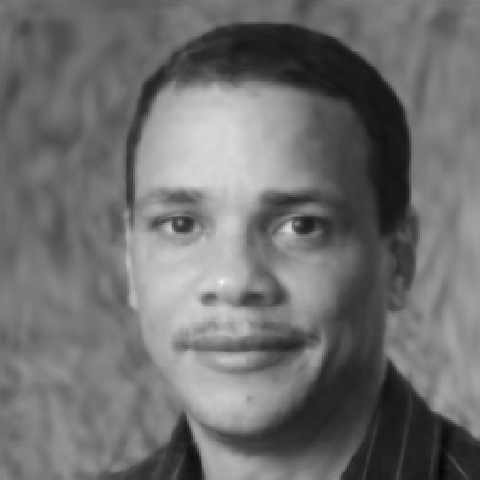} &         \includegraphics[width=\sza\columnwidth]{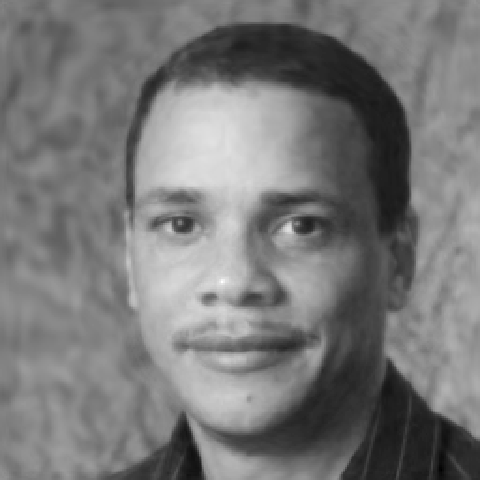} &     \includegraphics[width=\sza\columnwidth]{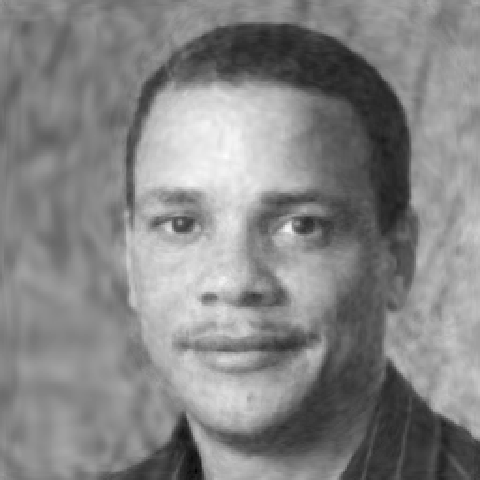} &     \includegraphics[width=\sza\columnwidth]{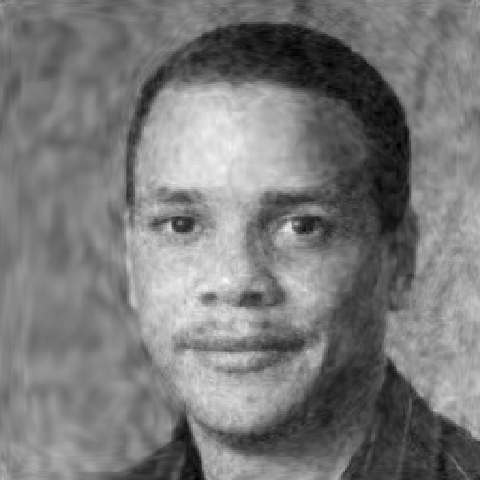} & 
    \hspace{0pt}
    \includegraphics[width=\sza\columnwidth]{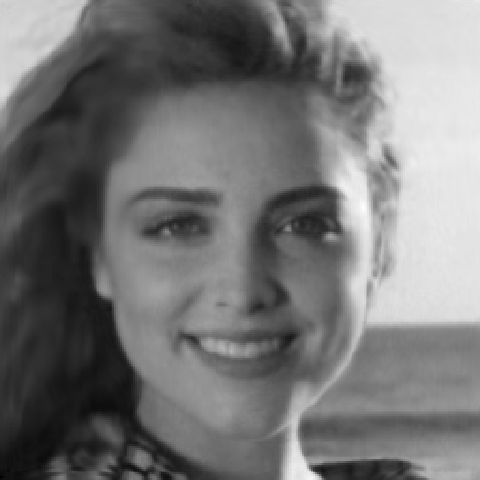} &         \includegraphics[width=\sza\columnwidth]{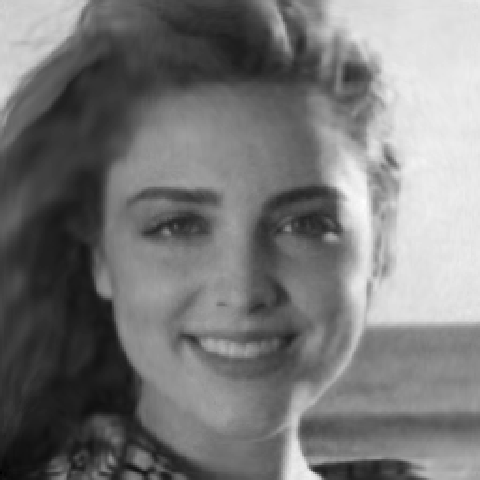} &     \includegraphics[width=\sza\columnwidth]{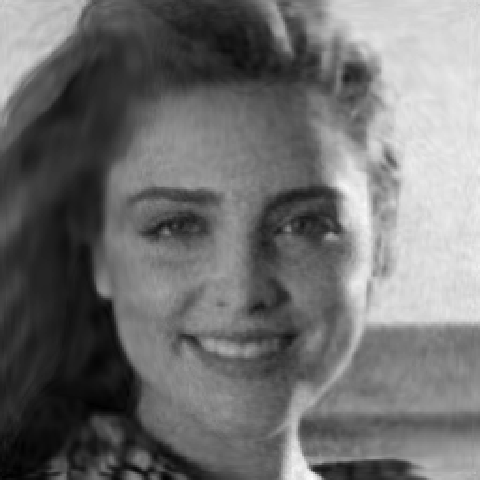} &     \includegraphics[width=\sza\columnwidth]{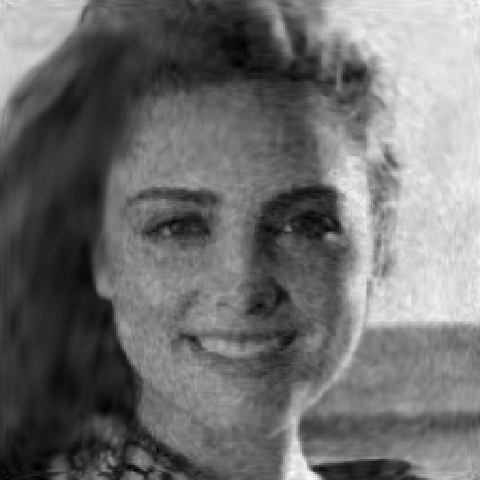}
        \\
    \rotatebox{90}{\hspace{0pt}\emph{\specialcell{\hspace{4pt} DL = 1e-4 \\ \hspace{4pt} TV = 3e-4}}}
    \hspace{1pt} &
    \includegraphics[width=\sza\columnwidth]{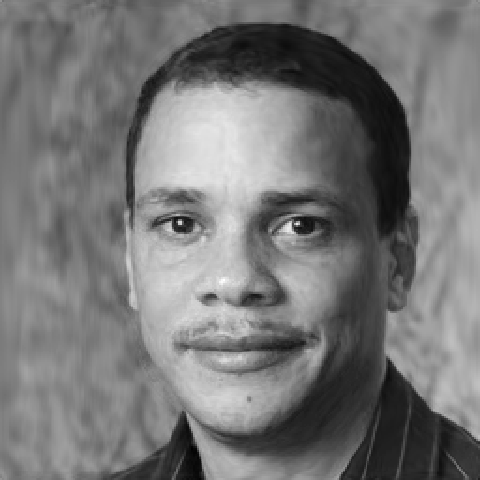} &          \includegraphics[width=\sza\columnwidth]{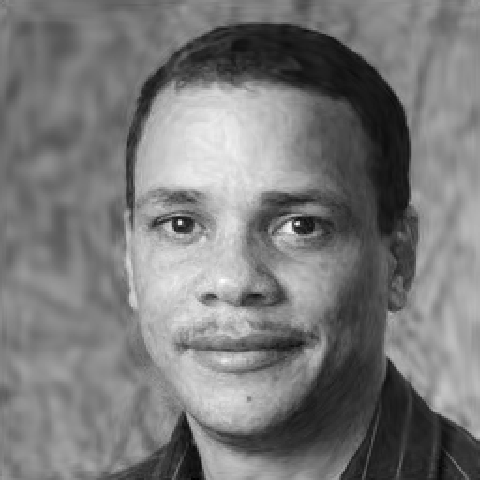} &      \includegraphics[width=\sza\columnwidth]{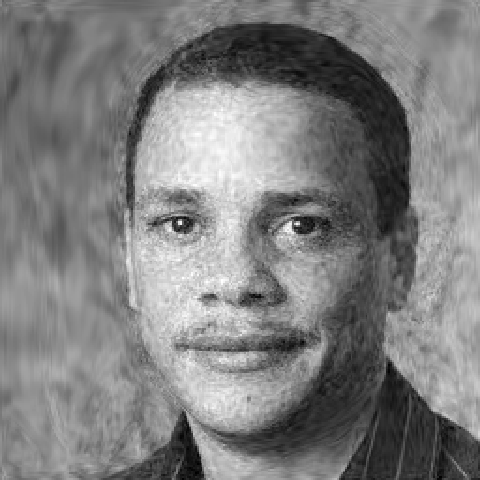} &      \includegraphics[width=\sza\columnwidth]{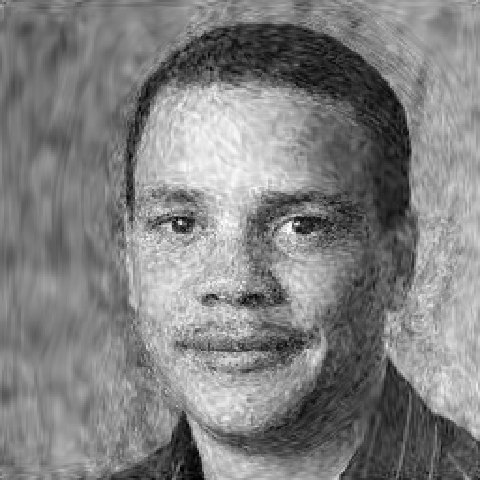} &  
    \hspace{0pt}
    \includegraphics[width=\sza\columnwidth]{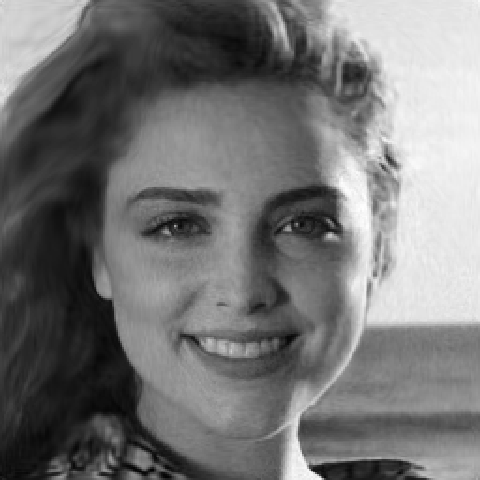} &          \includegraphics[width=\sza\columnwidth]{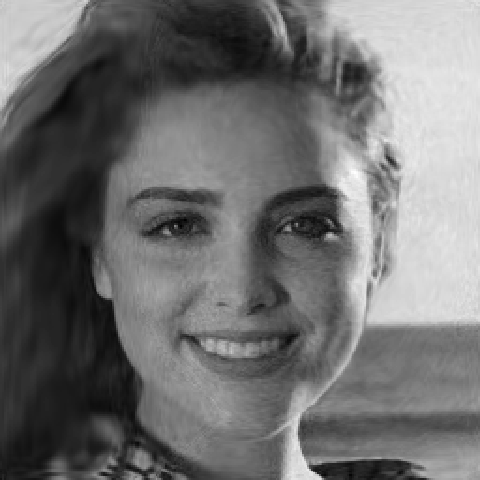} &      \includegraphics[width=\sza\columnwidth]{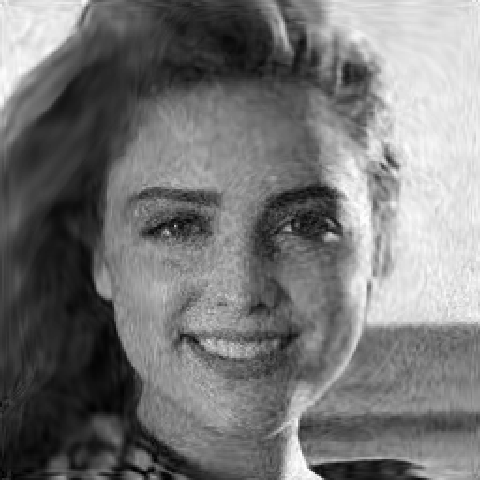} &      \includegraphics[width=\sza\columnwidth]{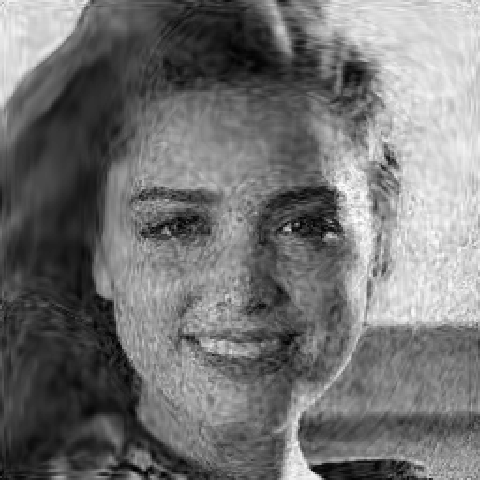} 
    \\
    \rotatebox{90}{\hspace{0pt}\emph{\specialcell{\hspace{4pt} DL = 1e-4 \\ \hspace{4pt} TV = 1e-3}}}
    \hspace{1pt} &
    \includegraphics[width=\sza\columnwidth]{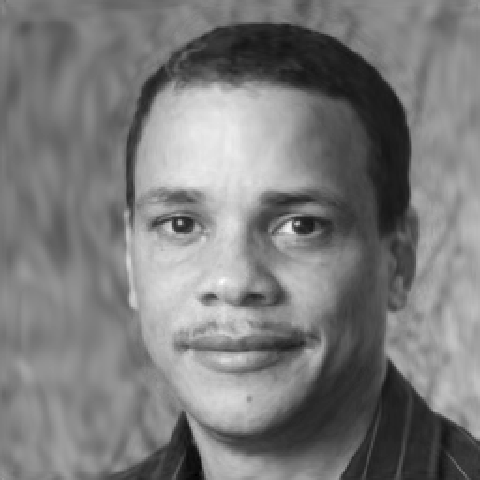} &          \includegraphics[width=\sza\columnwidth]{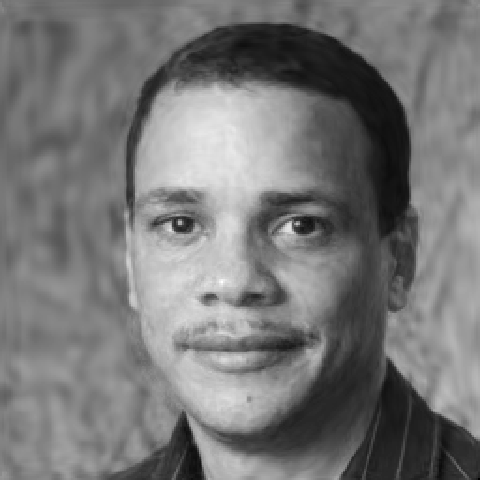} &       \includegraphics[width=\sza\columnwidth]{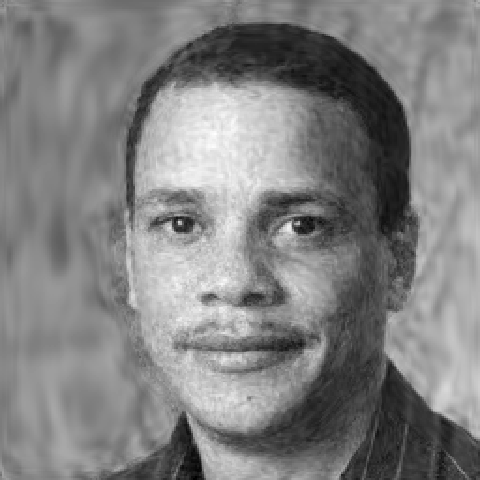} &       \includegraphics[width=\sza\columnwidth]{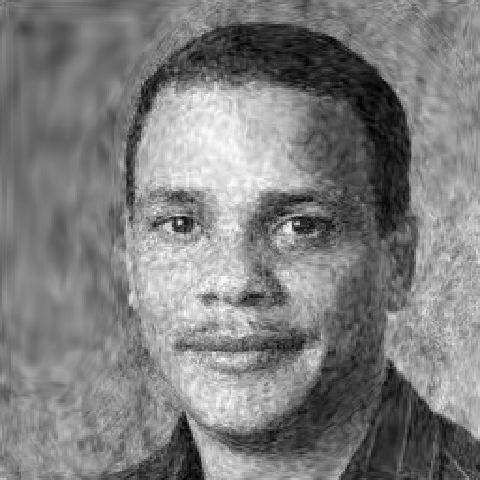} &    
    \hspace{0pt}
    \includegraphics[width=\sza\columnwidth]{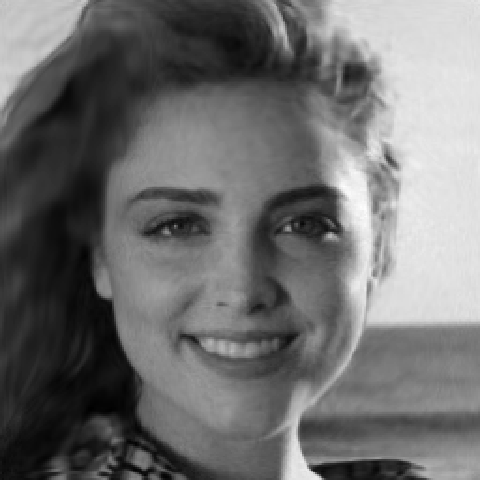} &          \includegraphics[width=\sza\columnwidth]{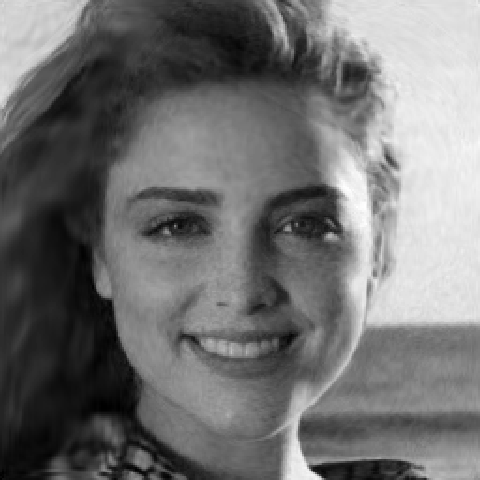} &       \includegraphics[width=\sza\columnwidth]{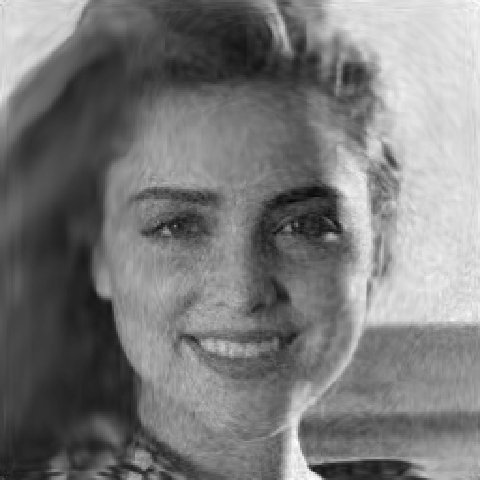} &       \includegraphics[width=\sza\columnwidth]{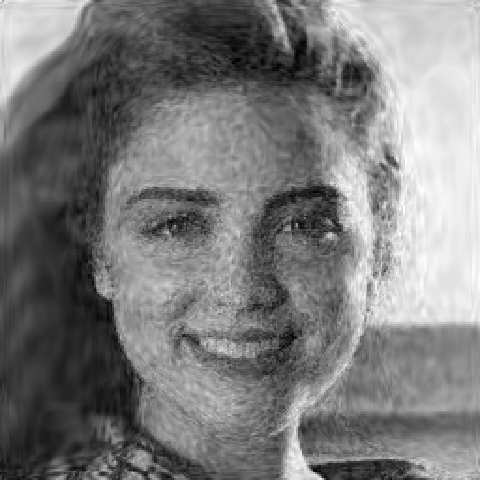}
    \\
    \rotatebox{90}{\hspace{0pt}\emph{\specialcell{\hspace{4pt} DL = 1e-4 \\ \hspace{4pt} TV = 3e-3}}}
    \hspace{1pt} &
    \includegraphics[width=\sza\columnwidth]{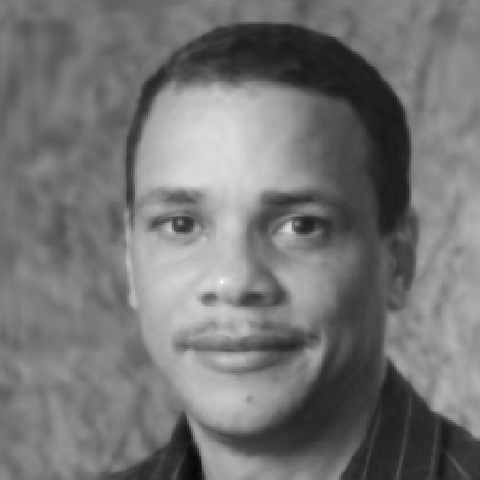} &          \includegraphics[width=\sza\columnwidth]{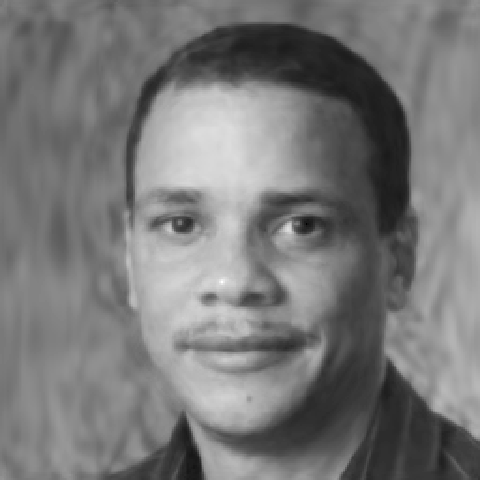} &        \includegraphics[width=\sza\columnwidth]{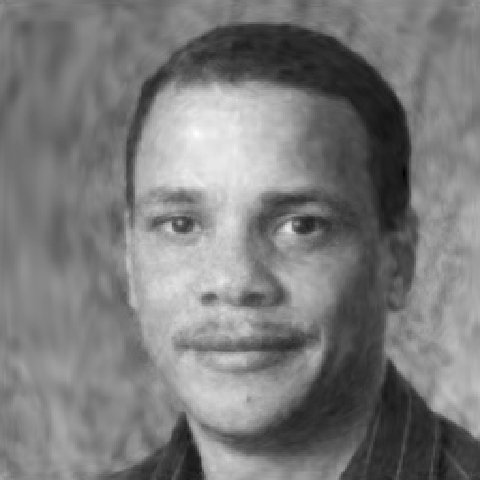} &        \includegraphics[width=\sza\columnwidth]{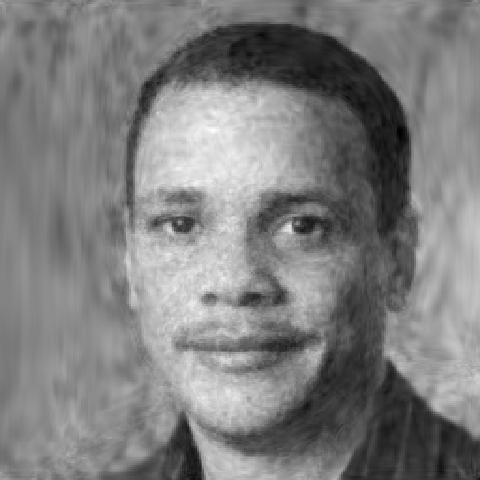} &    
    \hspace{0pt}
    \includegraphics[width=\sza\columnwidth]{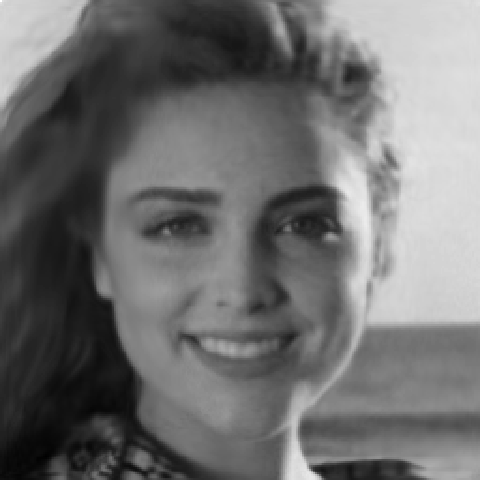} &          \includegraphics[width=\sza\columnwidth]{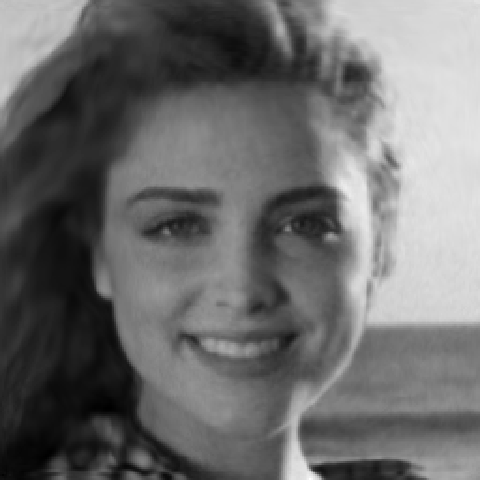} &        \includegraphics[width=\sza\columnwidth]{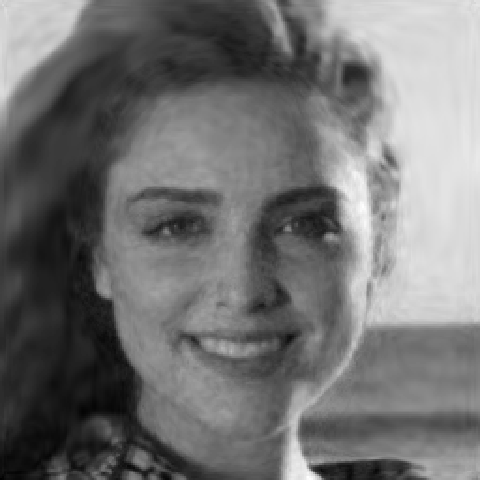} &        \includegraphics[width=\sza\columnwidth]{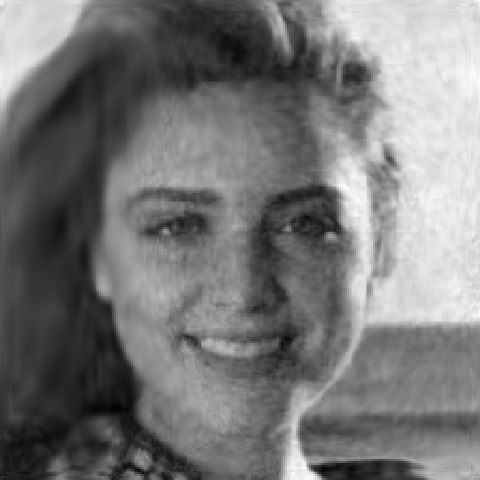} 
        \\
    \rotatebox{90}{\hspace{0pt}\emph{\specialcell{\hspace{4pt} DL = 1e-3 \\ \hspace{4pt} TV = 3e-4}}}
    \hspace{1pt} &
    \includegraphics[width=\sza\columnwidth]{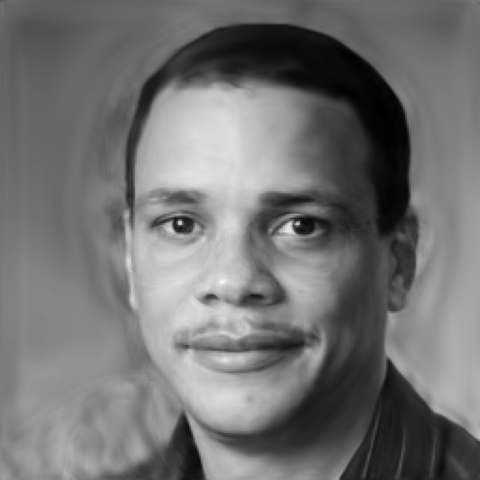} &          \includegraphics[width=\sza\columnwidth]{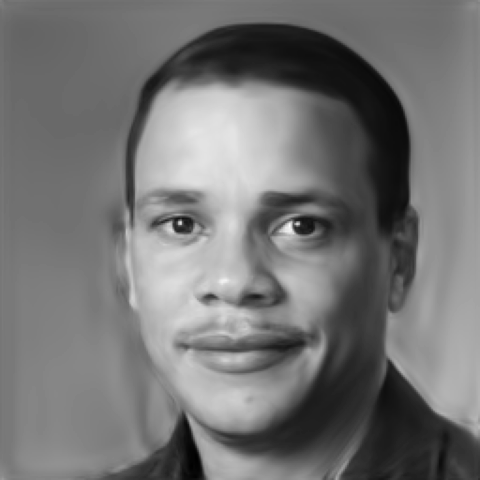} &     
    \includegraphics[width=\sza\columnwidth]{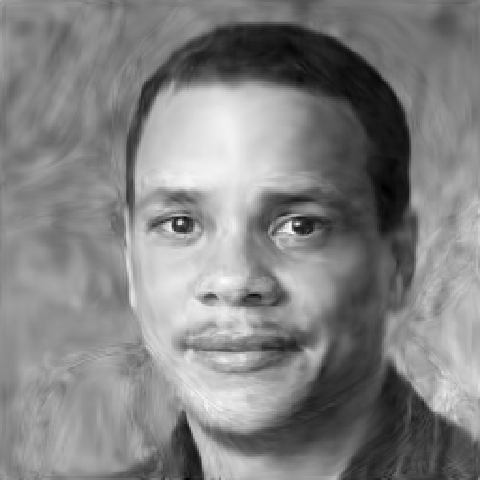} &     \includegraphics[width=\sza\columnwidth]{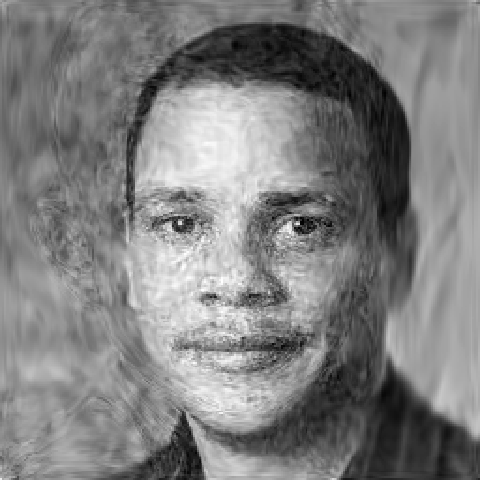} & 
    \hspace{0pt}
    \includegraphics[width=\sza\columnwidth]{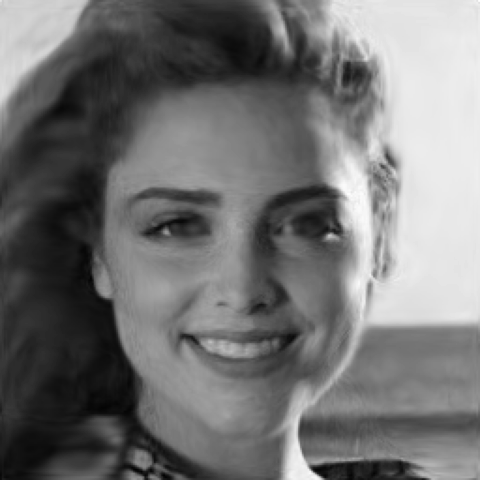} &          
    \includegraphics[width=\sza\columnwidth]{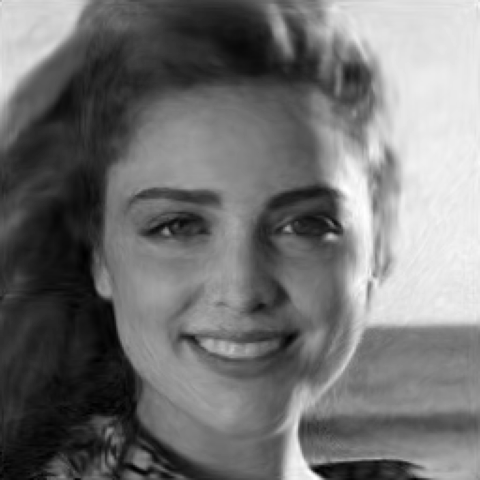} &
    \includegraphics[width=\sza\columnwidth]{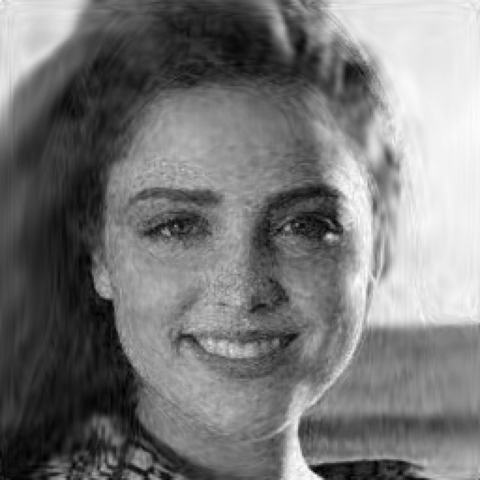} &
    \includegraphics[width=\sza\columnwidth]{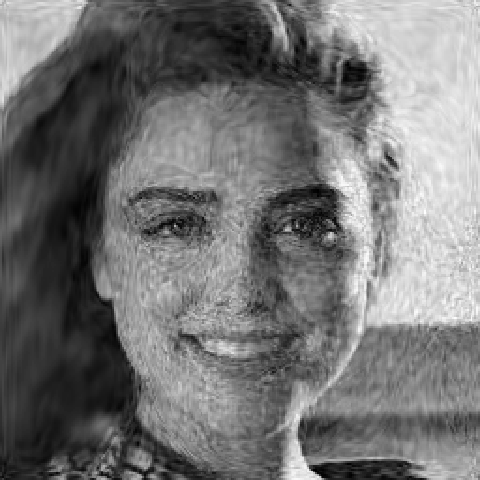}
    \\
    \rotatebox{90}{\hspace{0pt}\emph{\specialcell{\hspace{4pt} DL = 1e-3 \\ \hspace{4pt} TV = 1e-3}}}
    \hspace{1pt} &
    \includegraphics[width=\sza\columnwidth]{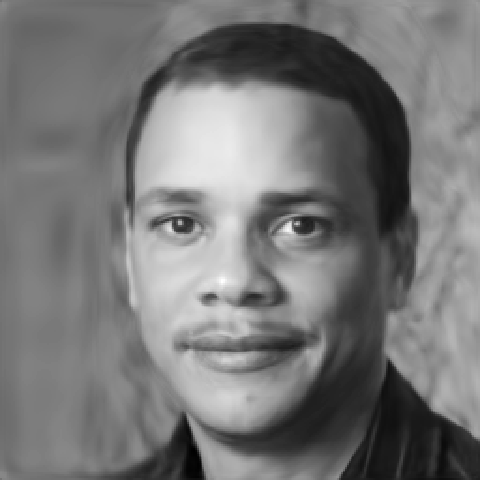} &          \includegraphics[width=\sza\columnwidth]{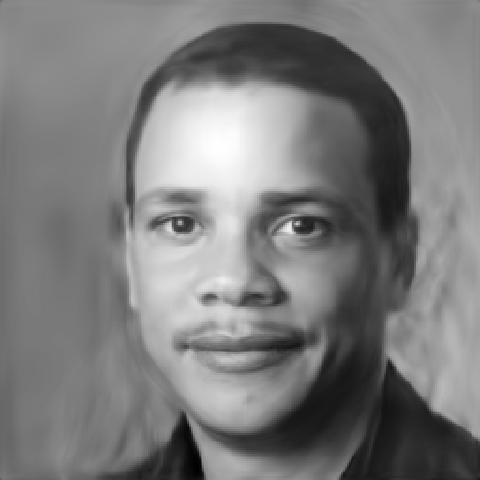} &     \includegraphics[width=\sza\columnwidth]{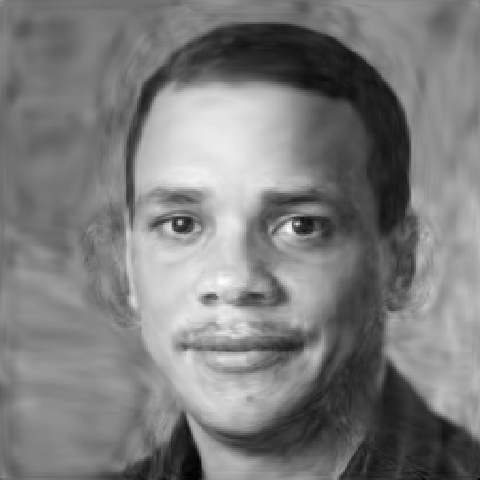} &     \includegraphics[width=\sza\columnwidth]{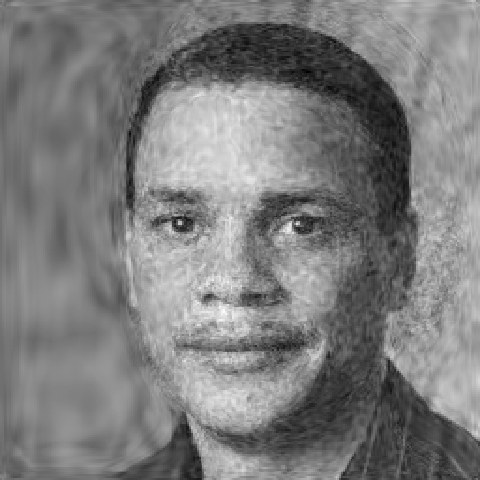} & 
    \hspace{0pt}
    \includegraphics[width=\sza\columnwidth]{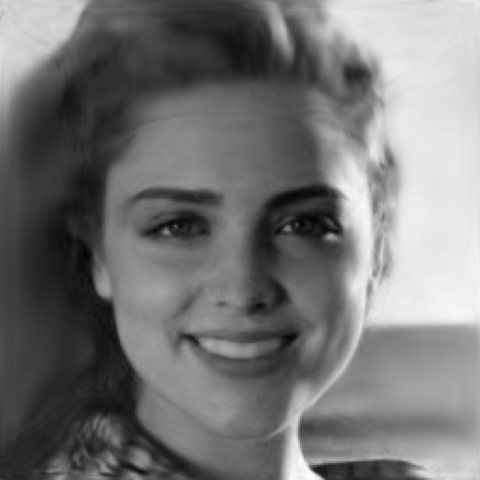} &          
    \includegraphics[width=\sza\columnwidth]{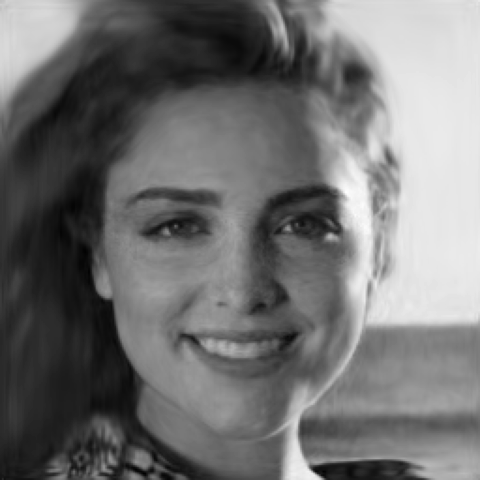} &
    \includegraphics[width=\sza\columnwidth]{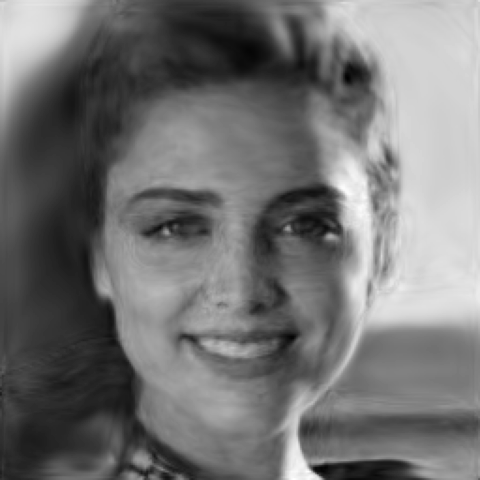} &
    \includegraphics[width=\sza\columnwidth]{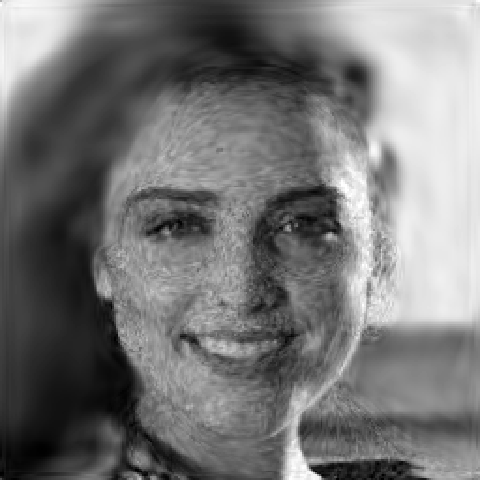}
    \\
    \rotatebox{90}{\hspace{0pt}\emph{\specialcell{\hspace{4pt} DL = 1e-3 \\ \hspace{4pt} TV = 3e-3}}}
    \hspace{1pt} &
    \includegraphics[width=\sza\columnwidth]{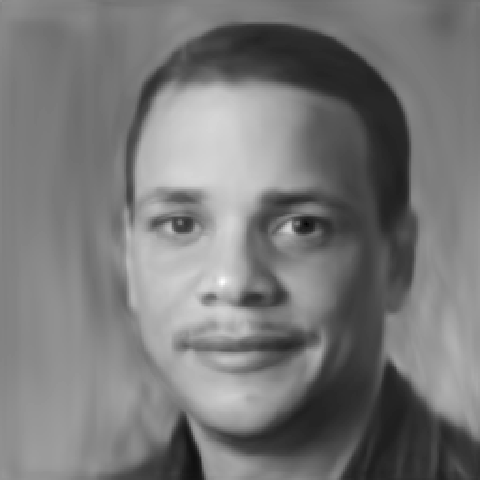} &          \includegraphics[width=\sza\columnwidth]{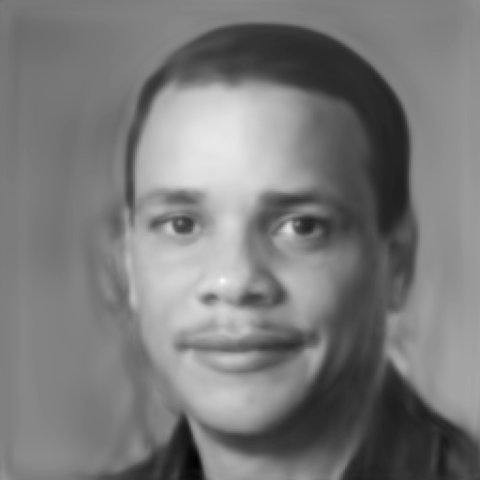} &   \includegraphics[width=\sza\columnwidth]{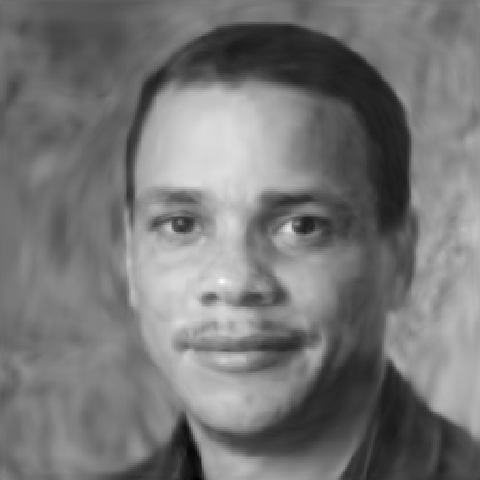} &   
    \includegraphics[width=\sza\columnwidth]{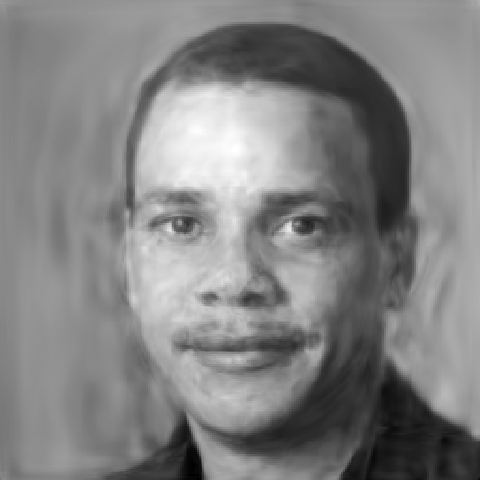} &  
    \hspace{0pt}
    \includegraphics[width=\sza\columnwidth]{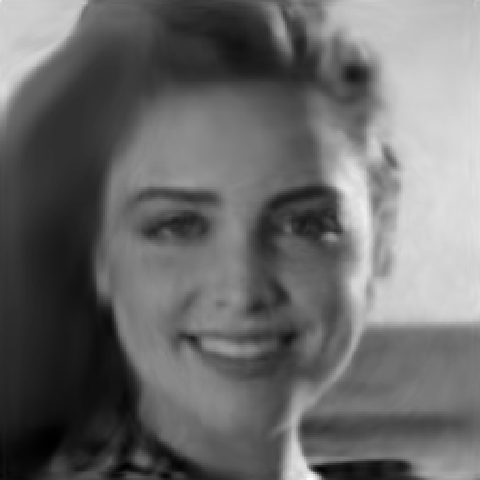} &          \includegraphics[width=\sza\columnwidth]{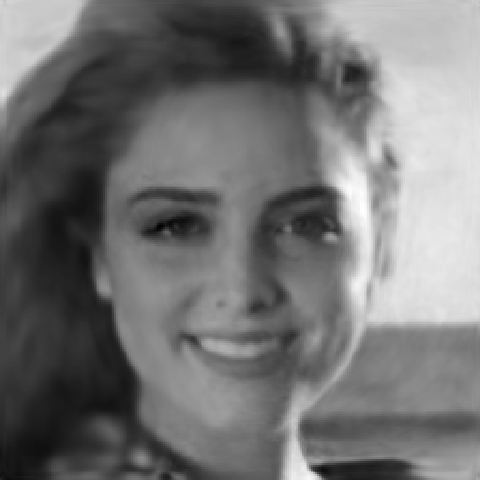} &     \includegraphics[width=\sza\columnwidth]{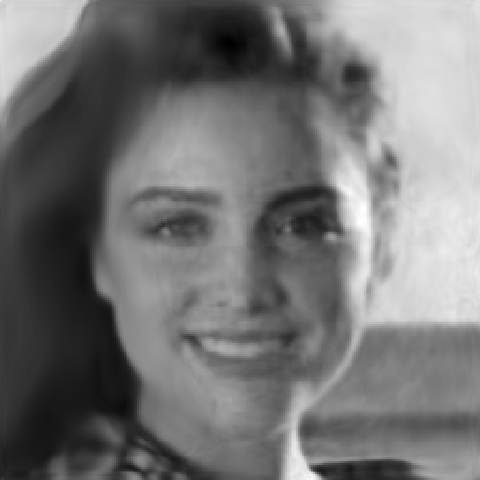} &     \includegraphics[width=\sza\columnwidth]{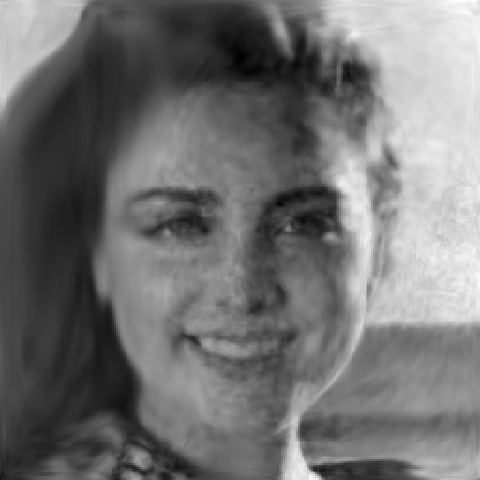}

\end{tabular}
\caption{Reconstructions from noisy data with varying noise levels and varying levels Discriminator Loss (DL) and Total Variation (TV) priors for 50\% overlap cases}
 \label{fig:rec3_noisy_priors}
\end{figure*}

For demonstration, different step sizes are utilized for various experiments resulting in different overlap percentages. The ptychographic step sizes for the probe movement are chosen to be 8, 16, 24, 32, and 48 pixels, which correspond to 75\%, 50\%, 25\%, 0\%, and -50\% overlaps. Most of the experiments for comparison are done for 50\% overlap cases to show the limitations in a high overlap percentage, but reconstructions for no-overlap situations are done to show the effectiveness of the algorithm. Four different Poisson noise levels are used for comparison, which are denoted  by $\sigma$ values showing their relative intensity compared to the peak intensity values, and these are chosen to be 0.2, 0.5, 2, and 5.

\subsection{Reconstructions and Comparison}

To evaluate the robustness of the proposed approach, we show the reconstructions for different face images, different noise levels, various adjustments, and changing priors. We also present methods in the literature to be compared to the proposed approach, present the reconstructions for the simulations for all methods, and discuss the results qualitatively and quantitatively.

The reconstructions are done in PyTorch with the proposed algorithm by using the minimization shown Eq.~\ref{Eq:converted}. In each experiment, a different aspect of the algorithm is tested with varying parameters. For experiments optimizing the input latent vector, standard gradient descent with a learning rate between $2\times 10^{-6}$ and $2\times 10^{-5}$ for 1000 epochs is used to solve the minimization in Eq.~\ref{Eq:L1_dgp} by keeping the weights of the network constant. For experiments optimizing the network weights without progressive adjustment, the expression in Eq.~\ref{Eq:converted} is minimized after initial adjustment  of the latent vector using Adam optimizer \cite{kingma2014method} with learning rates between $2\times 10^{-5}$ and $2\times 10^{-4}$ is used till convergence. In the cases where progressive adjustment is applied, the update of the weights starts from the last  layer and consequent layers are unfrozen at each $800^{th}$ step. At the $5^{th}$ step, all layers are started to be updated.

In Fig~\ref{fig:rec1_nf}, reconstructions for three face images can be seen compared to the ground truth image without TV and discriminator priors. Reconstructions in the second column show that adjustment of the latent vector is not enough for a successful reconstruction.  This is due to the limited representation capability of the generator network showing that deep generative priors (DGPs) are not enough for this problem by itself. However, the reconstructions after the adjustment of the network weights are similar to the ground truth images showing the capability of deep image priors (DIPs). On the other hand, although the reconstructions are very good in the facial parts, the background is not as close to the ground truth. That is due to the lack of DGPs in these parts. 

We show that the reconstructions with the proposed method are successful. However, we suggested that applying progressive adjustment of the network weights would increase the reconstruction quality. In Fig.~\ref{fig:rec2_prog_nf}, the reconstructions for 50\% probe overlap with noise-free data is shown. Here, it can be observed that optimizing the layers while freezing some of them gives reconstructions that are far from the ideal solution. Thus, adjustment of all layers is required for successful outputs. However, it is a highly convex minimization problem, and adjustment of layers progressively gives a significantly better starting point for the final step. This can be observed in the fifth column of Fig.~\ref{fig:rec2_prog_nf}, where we can see that the background and the general shape of the heads look similar to the initial steps in the reconstructions; however, the facial expressions become more similar to the required solution. That results in an improved solution compared to the non-progressive reconstruction.

\begin{figure*}[!ht]
    \centering
    \setlength{\tabcolsep}{\szb}

    \begin{tabular}{ccccccc}
    & {ePIE} & {Proposed} & {GT} 
    & {ePIE} & {Proposed} & {GT} 
    \vspace{2pt}
    \\
    \rotatebox{90}{\hspace{12pt}\emph{$75\%$}}
    \hspace{1pt} &
    \includegraphics[width=\sza\columnwidth]{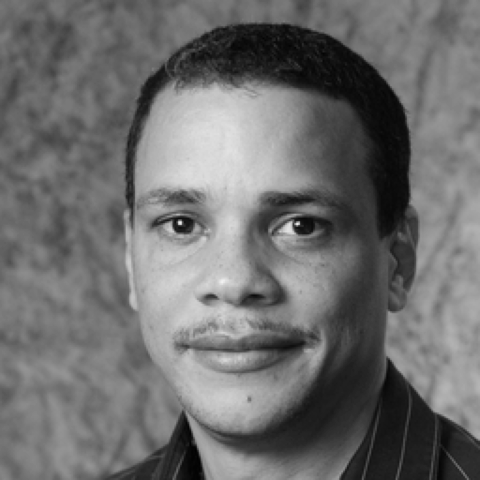} & 
    \includegraphics[width=\sza\columnwidth]{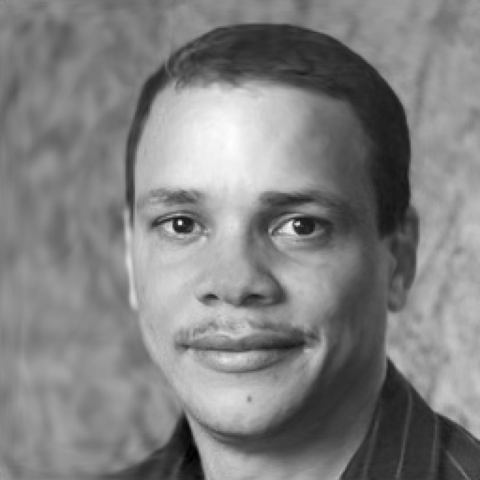} &
    \includegraphics[width=\sza\columnwidth]{figs/gt1.png} &
    \hspace{2pt}
    \includegraphics[width=\sza\columnwidth]{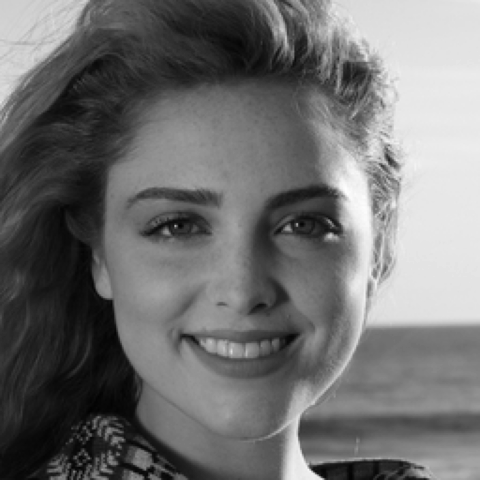} & 
    \includegraphics[width=\sza\columnwidth]{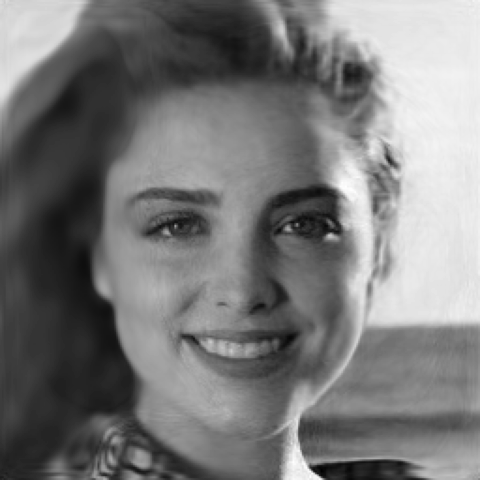} &
    \includegraphics[width=\sza\columnwidth]{figs/gt2.png}
    \\
    \rotatebox{90}{\hspace{12pt}\emph{$50\%$}}
    \hspace{1pt} &
    \includegraphics[width=\sza\columnwidth]{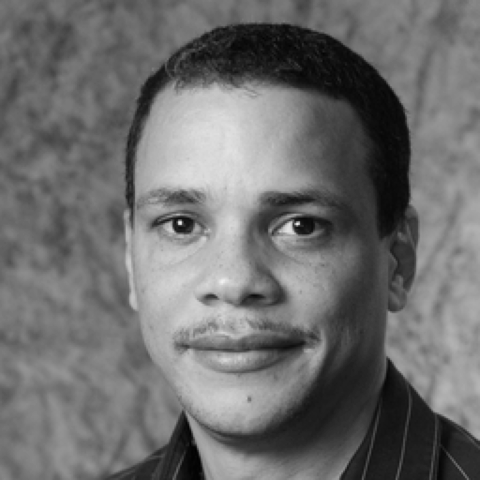} & 
    \includegraphics[width=\sza\columnwidth]{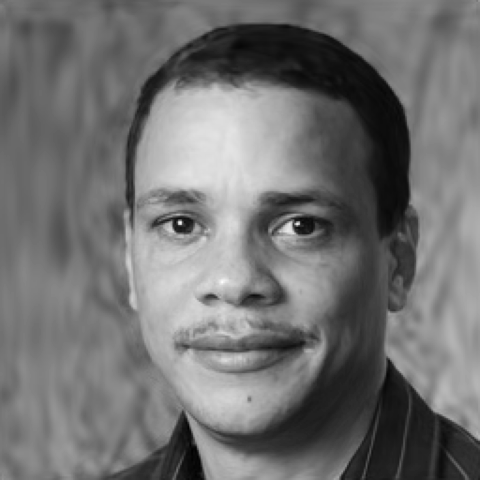} &
    \includegraphics[width=\sza\columnwidth]{figs/gt1.png} &
    \hspace{2pt}
    \includegraphics[width=\sza\columnwidth]{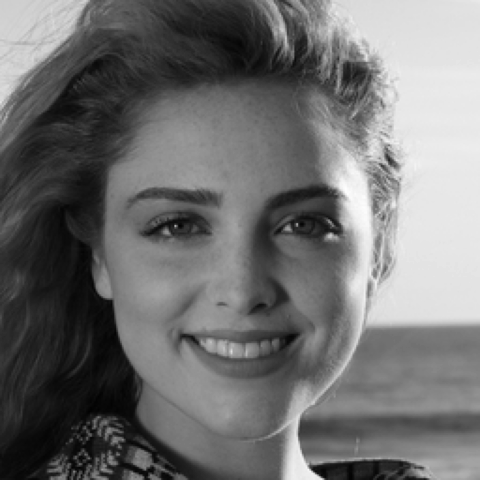} & 
    \includegraphics[width=\sza\columnwidth]{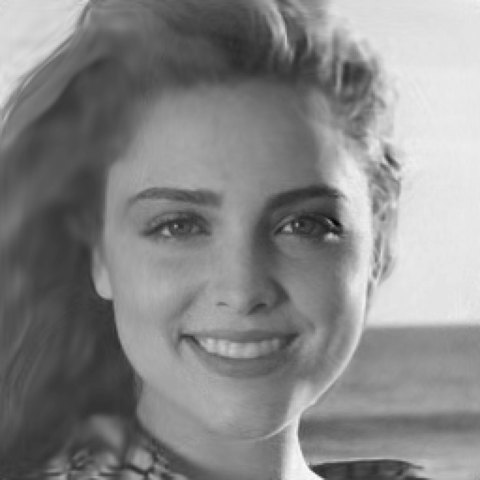} &
    \includegraphics[width=\sza\columnwidth]{figs/gt2.png}
    \\
    \rotatebox{90}{\hspace{12pt}\emph{$25\%$}}
    \hspace{1pt} &
    \includegraphics[width=\sza\columnwidth]{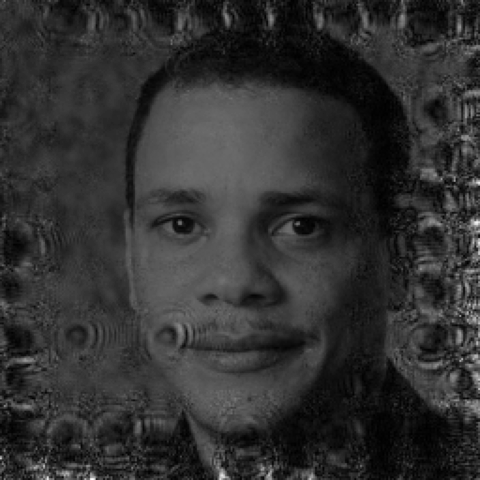} & 
    \includegraphics[width=\sza\columnwidth]{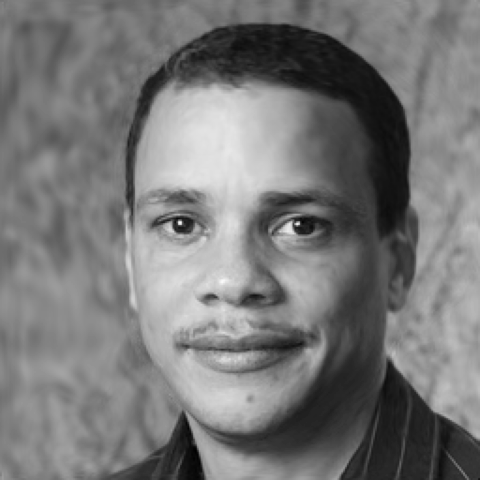} &
    \includegraphics[width=\sza\columnwidth]{figs/gt1.png} &
    \hspace{2pt}
    \includegraphics[width=\sza\columnwidth]{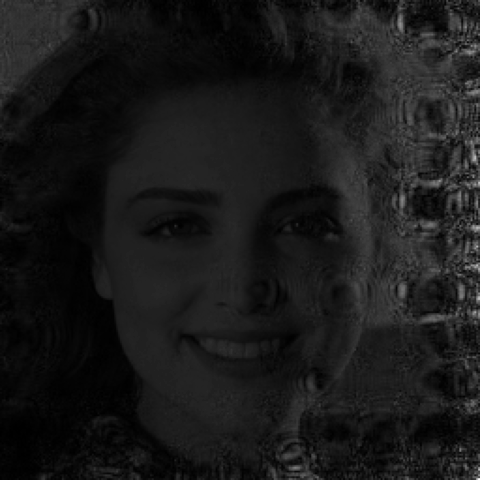} & 
    \includegraphics[width=\sza\columnwidth]{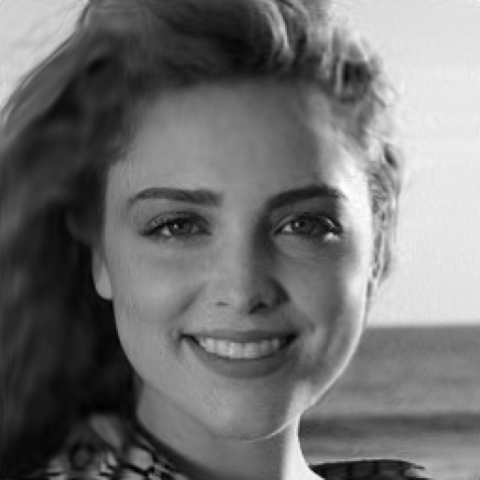} &
    \includegraphics[width=\sza\columnwidth]{figs/gt2.png}
    \\
    \rotatebox{90}{\hspace{12pt}\emph{$0\%$}}
    \hspace{1pt} &
    \includegraphics[width=\sza\columnwidth]{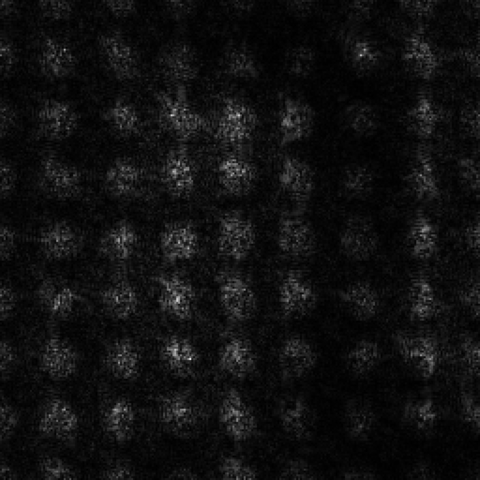} & 
    \includegraphics[width=\sza\columnwidth]{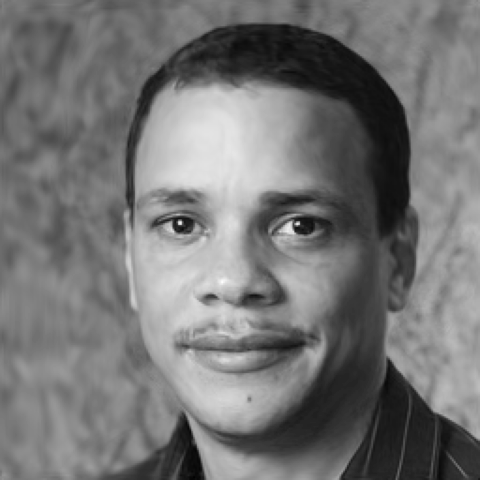} &
    \includegraphics[width=\sza\columnwidth]{figs/gt1.png} &
    \hspace{2pt}
    \includegraphics[width=\sza\columnwidth]{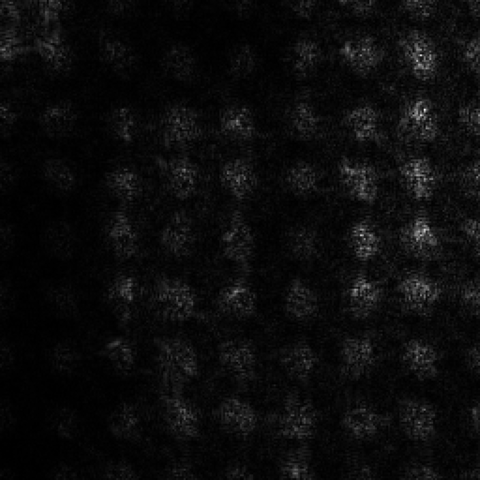} & 
    \includegraphics[width=\sza\columnwidth]{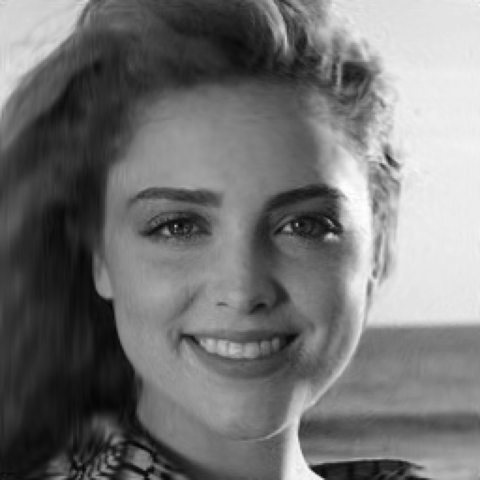} &
    \includegraphics[width=\sza\columnwidth]{figs/gt2.png}
    \\
    \rotatebox{90}{\hspace{12pt}\emph{$-50\%$}}
    \hspace{1pt} &
    \includegraphics[width=\sza\columnwidth]{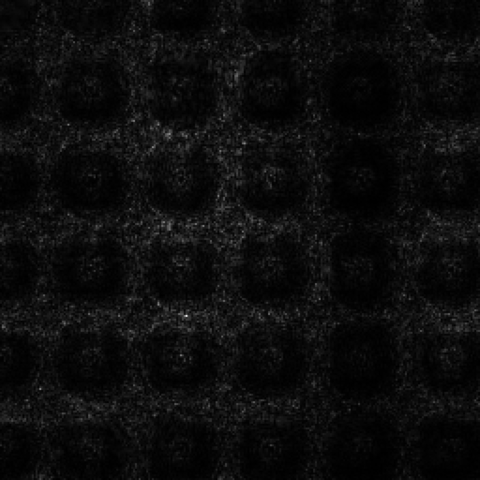} & 
    \includegraphics[width=\sza\columnwidth]{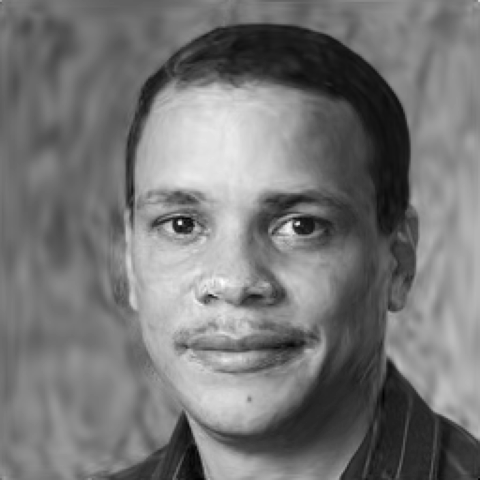} &
    \includegraphics[width=\sza\columnwidth]{figs/gt1.png} &
    \hspace{2pt}
    \includegraphics[width=\sza\columnwidth]{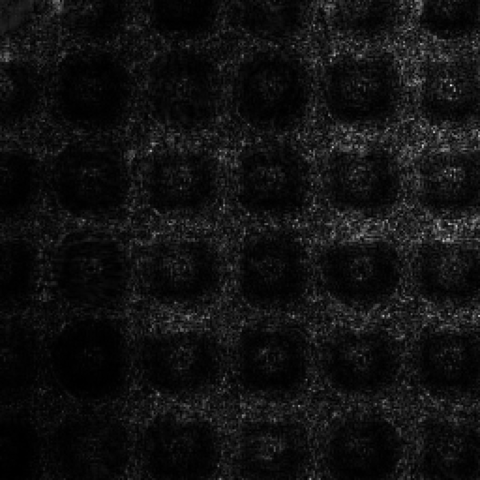} & 
    \includegraphics[width=\sza\columnwidth]{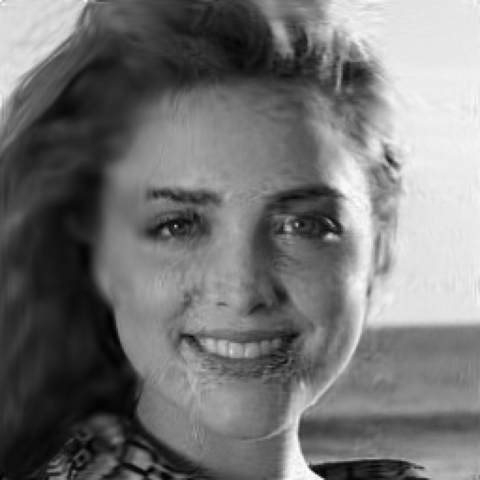} &
    \includegraphics[width=\sza\columnwidth]{figs/gt2.png}

\end{tabular}
\caption{Comparison of the reconstructions from the proposed method against ePIE and ground truth for noise-free case for different overlap percentages}
 \label{fig:overlap_nf}
\end{figure*}
\setlength\tabcolsep{3.2pt}
\begin{table}[!b]
\footnotesize
\caption{Structural Similarity Index (SSIM) for different noise levels and different amounts of regularization.}
\centering
\begin{tabular}{@{}c|c|c|c|c||c|c|c|c|@{}}
\multicolumn{1}{|c|}{\textbf{SSIM}} &
\multicolumn{4}{c||}{\textbf{Image 1}} & \multicolumn{4}{c|}{\textbf{Image 2}} \\
% \cmidrule(l){2-9} 
\midrule
\multicolumn{1}{|c|}{Noise Level ($\sigma^2$)} & {$0.2$} & {$0.5$} & {$2$} & {$5$} & {$0.2$} & {$0.5$} & {$2$} & {$5$} \\ \midrule
\multicolumn{1}{|c|}{DL(0), TV(0)} & \textbf{0.96} & 0.94 & 0.86 & 0.75 & 0.94 & 0.88 & 0.85 & 0.76 \\
\multicolumn{1}{|c|}{DL(1e-5), TV(3e-4)} & \textbf{0.97} & \textbf{0.95}  & 0.86 & 0.77 & \textbf{0.95} & \textbf{0.95} & 0.86 & 0.75 \\
\multicolumn{1}{|c|}{DL(1e-5), TV(1e-3)} & \textbf{0.96} & \textbf{0.96} & 0.89 & 0.81 & \textbf{0.95} & \textbf{0.93} & 0.89 & 0.81 \\
\multicolumn{1}{|c|}{DL(1e-5), TV(3e-3)} & 0.94 & 0.94 & 0.89 & \textbf{0.87} & 0.92 & 0.91 & \textbf{0.91} & \textbf{0.86} \\
\multicolumn{1}{|c|}{DL(1e-4), TV(3e-4)} & 0.96 & \textbf{0.95} & 0.86 & 0.75 & \textbf{0.97} & \textbf{0.93} & 0.87 & 0.77 \\
\multicolumn{1}{|c|}{DL(1e-4), TV(1e-3)} & 0.94 & \textbf{0.95} & 0.88 & 0.78 & \textbf{0.95} & \textbf{0.95} & 0.86 & 0.79 \\
\multicolumn{1}{|c|}{DL(1e-4), TV(3e-3)} & 0.94 & 0.92 & \textbf{0.91} & \textbf{0.86} & 0.92 & 0.92 & \textbf{0.90} & \textbf{0.86} \\
\multicolumn{1}{|c|}{DL(1e-3), TV(3e-4)} & 0.92 & 0.90 & \textbf{0.90} & 0.81 & 0.93 & \textbf{0.93} & 0.88 & 0.78 \\
\multicolumn{1}{|c|}{DL(1e-3), TV(1e-3)} & 0.92 & 0.88 & \textbf{0.90} & 0.81 & 0.87 & 0.92 & 0.87 & 0.83 \\
\multicolumn{1}{|c|}{DL(1e-3), TV(3e-3)} & 0.87 & 0.87 & \textbf{0.90} & \textbf{0.86} & 0.88 & 0.89 & 0.88 & \textbf{0.85} \\
\hline
\end{tabular}
\label{tab:tab1}
\end{table}

\setlength\tabcolsep{3.2pt}
\begin{table}[hb]
\footnotesize
\caption{Structural Similarity Index (SSIM) comparison for ePIE and proposed (Prop) methods and different overlap percentages using noise-free data.}
\centering
\begin{tabular}{@{}c|c|c||c|c|@{}}
\multicolumn{1}{|c|}{\textbf{SSIM}} &
\multicolumn{2}{c||}{\textbf{Image 1}} & \multicolumn{2}{c|}{\textbf{Image 2}} \\
% \cmidrule(l){2-9} 
\midrule
\multicolumn{1}{|c|}{Method} & {ePIE} & {Prop} & {ePIE} & {Prop} \\ \midrule
\multicolumn{1}{|c|}{Overlap = $75\%$} & \textbf{1.00} & 0.98 & \textbf{1.00} & 0.98 \\
\multicolumn{1}{|c|}{Overlap = $50\%$} & \textbf{0.99} & 0.98 & \textbf{0.99} & 0.97 \\
\multicolumn{1}{|c|}{Overlap = $25\%$} & 0.54 & \textbf{0.97} & 0.33 & \textbf{0.96} \\
\multicolumn{1}{|c|}{Overlap = $0\%$} & 0.16 & \textbf{0.94} & 0.14 & \textbf{0.95} \\
\multicolumn{1}{|c|}{Overlap = $-50\%$} & 0.15 & \textbf{0.93} & 0.18 & \textbf{0.93} \\
\hline
\end{tabular}
\label{tab:tab2}
\end{table}

The noise in the diffraction data for x-ray ptychography creates some artifacts in the reconstructions. For the proposed method, these artifacts are affected in two directions. DGPs and DIPs used in the proposed method provide some additional information, leading to a constrained solution avoiding the noise. On the other hand, since the loss function is minimized designed to fit the diffraction images acquired from the reconstructed data to the input data, the network forces the reconstructions to be noisy as well. Thus, we add two different regularizers: discriminator loss and total variation as shown in Eq.~\ref{Eq:converted}. Reconstructions are shown for varying $\lambda$ levels of regularizers for different noise levels in Fig~\ref{fig:rec3_noisy_priors}. For low noise levels, the algorithm performs well without a need for priors. However, the reconstructions show significant artifacts with the increased noise in diffraction data. 

The structural similarity index (SSIM) scores for Fig~\ref{fig:rec3_noisy_priors} are given in Table~\ref{tab:tab1}. If we investigate the effect of regularization in both of them, we can see that the addition of discriminator loss as a regularizer increases the effect of DGPs, forcing the output not to diverge from the face domain. Although it seems like this improves the reconstruction quality, it might cause divergence from the data itself, as can be seen from various cases in the bottom rows. In addition, discriminator loss forces the reconstruction to an "average face" from the learned domain, which means that it inherently forces smoothness. This effect can be seen to be helpful for high noise scenarios, resulting in higher SSIM scores with increased discriminator loss regularization effect. However, it also reduces the fit to the data itself, resulting in over-smoothed reconstructions in low noise scenarios.

Even though the goals of using total variation regularization and discriminator loss regularization are similar in terms of smoothing, the effects are significantly different from each other. Forcing the solution to be close to averaged out face prior does not necessarily enforce piece-wise smoothness as total variation regularization does. In Fig~\ref{fig:rec3_noisy_priors}, it can be observed that an increased multiplier of TV regularizer results in smoother areas in specific parts of the face such as the forehead or cheeks. However, since the face images are not piece-wise smooth in most cases, the effect of the regularization is lower than it could have been in other datasets. In addition, unlike the effect of discriminator loss regularization, the effect of total variation regularization is more linear and consistent, which results in a gradual increase or decrease in reconstruction quality. Similarly, the SSIM values in Table~\ref{tab:tab1} are highest for the higher multiplier of TV regularizer for high noise levels and highest for the lower multiplier for low noise levels.

The most crucial advantage of our proposed method is the performance when there is a low overlap between the scanning positions in x-ray ptychography. The constraints that deep generative priors provide are crucial for having a good initial reconstruction, and the optimization that deep image priors drive the solution towards the ground truth. To test how effective these optimizations are, experiments with different overlap percentages are done with the standard Extended Ptychographic Iterative Engine (ePIE) method, and a comparison with the proposed method is made.

\begin{figure*}[!ht]
    \centering
    \setlength{\tabcolsep}{\szb}

    \begin{tabular}{ccccccccc}
    & {ePIE} & \begin{tabular}{@{}c@{}}Proposed \\ w/out priors\end{tabular} & \begin{tabular}{@{}c@{}}Proposed \\ w/ priors\end{tabular} & {GT}
    & {ePIE} & \begin{tabular}{@{}c@{}}Proposed \\ w/out priors\end{tabular} & \begin{tabular}{@{}c@{}}Proposed \\ w/ priors\end{tabular} & {GT} 
    \vspace{2pt}
    \\
    \rotatebox{90}{\hspace{12pt}\emph{$75\%$}}
    \hspace{1pt} &
    \includegraphics[width=\sza\columnwidth]{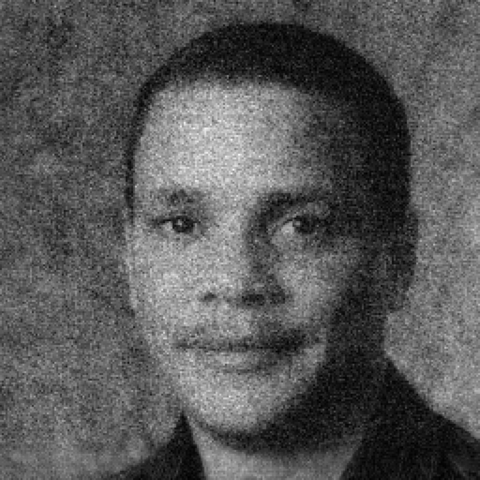} &    
    \includegraphics[width=\sza\columnwidth]{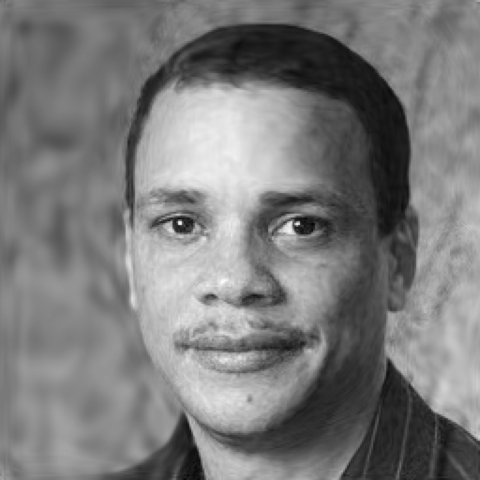} &
    \includegraphics[width=\sza\columnwidth]{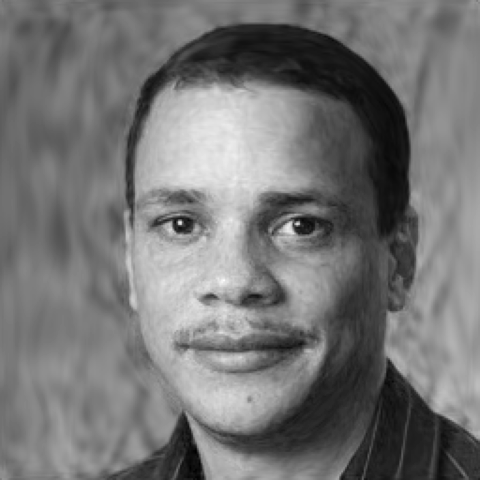} &
    \includegraphics[width=\sza\columnwidth]{figs/gt1.png} &
    \hspace{2pt}
    \includegraphics[width=\sza\columnwidth]{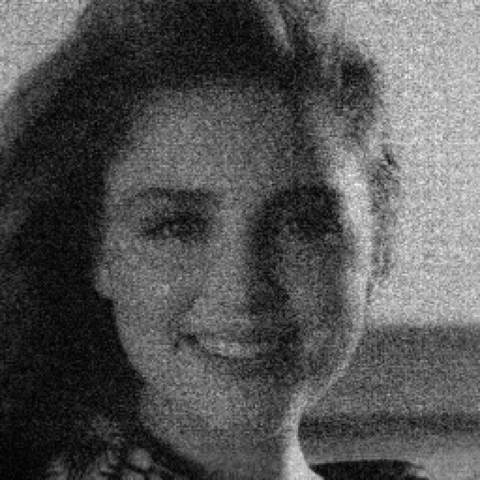} &       
    \includegraphics[width=\sza\columnwidth]{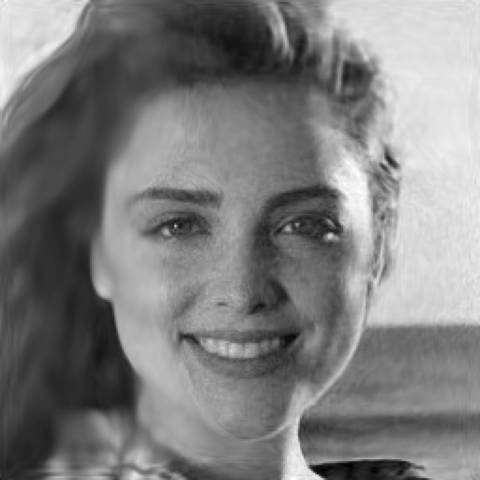} &
    \includegraphics[width=\sza\columnwidth]{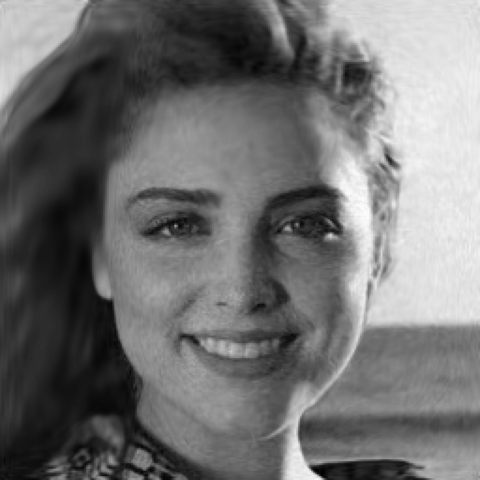} &
    \includegraphics[width=\sza\columnwidth]{figs/gt2.png}
    \\
    \rotatebox{90}{\hspace{12pt}\emph{$50\%$}}
    \hspace{1pt} &
    \includegraphics[width=\sza\columnwidth]{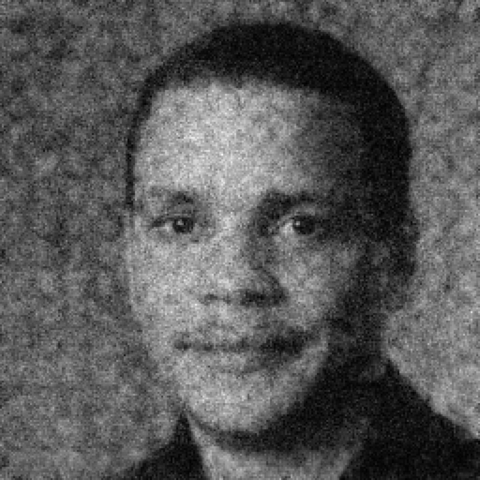} &        \includegraphics[width=\sza\columnwidth]{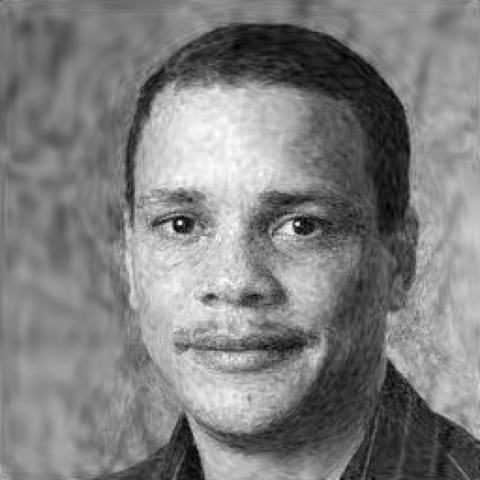} &
    \includegraphics[width=\sza\columnwidth]{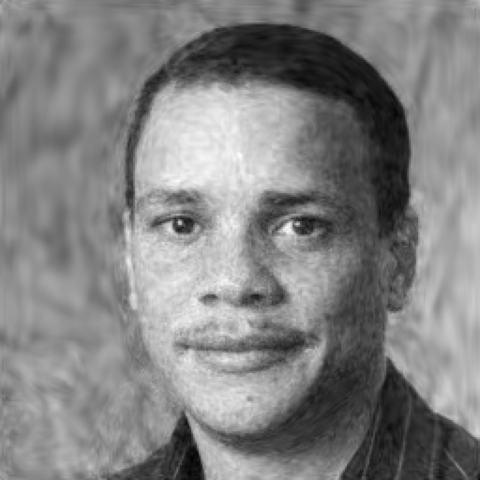} &
    \includegraphics[width=\sza\columnwidth]{figs/gt1.png} & 
    \hspace{2pt}
    \includegraphics[width=\sza\columnwidth]{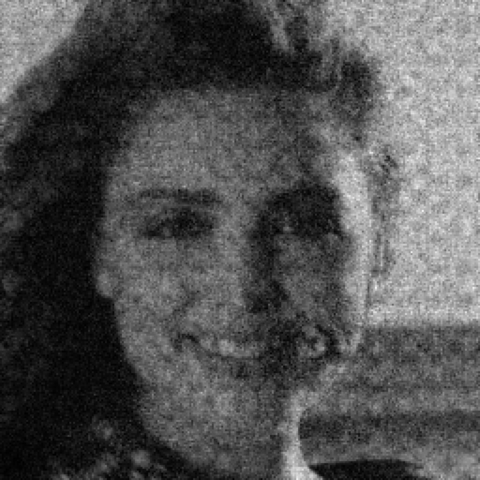} &    
    \includegraphics[width=\sza\columnwidth]{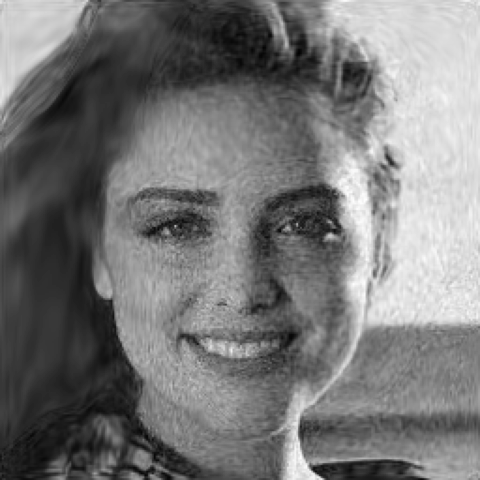} &
    \includegraphics[width=\sza\columnwidth]{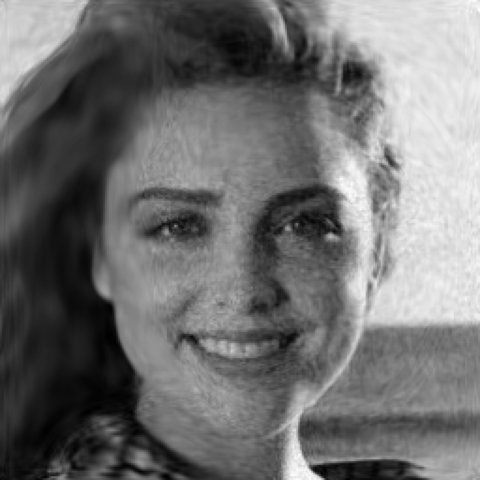} &
    \includegraphics[width=\sza\columnwidth]{figs/gt2.png}
    \\
    \rotatebox{90}{\hspace{12pt}\emph{$25\%$}}
    \hspace{1pt} &
    \includegraphics[width=\sza\columnwidth]{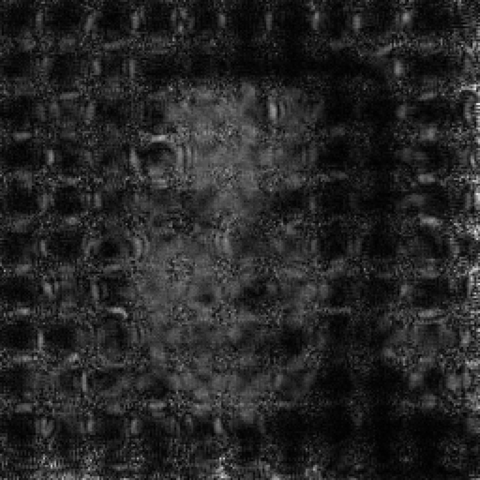} &  
    \includegraphics[width=\sza\columnwidth]{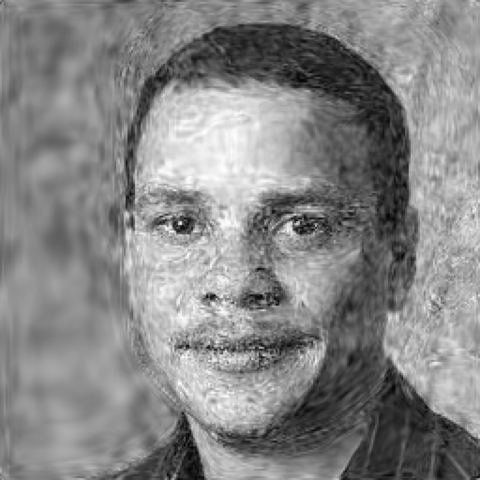} &
    \includegraphics[width=\sza\columnwidth]{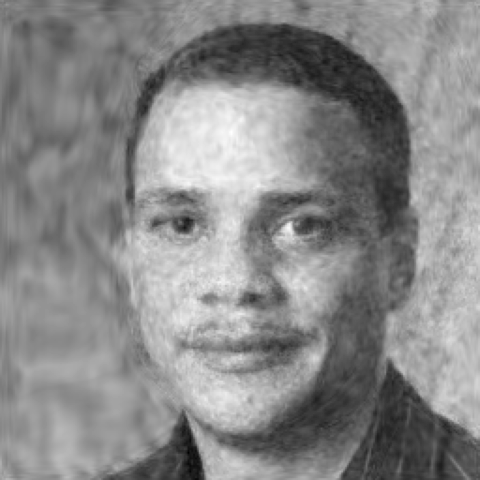} &
    \includegraphics[width=\sza\columnwidth]{figs/gt1.png} &
    \hspace{2pt}
    \includegraphics[width=\sza\columnwidth]{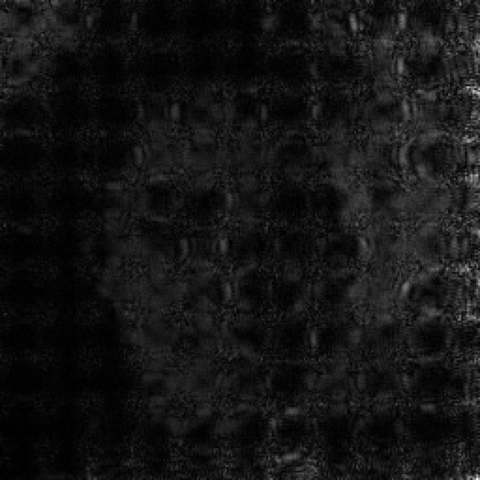} &      
    \includegraphics[width=\sza\columnwidth]{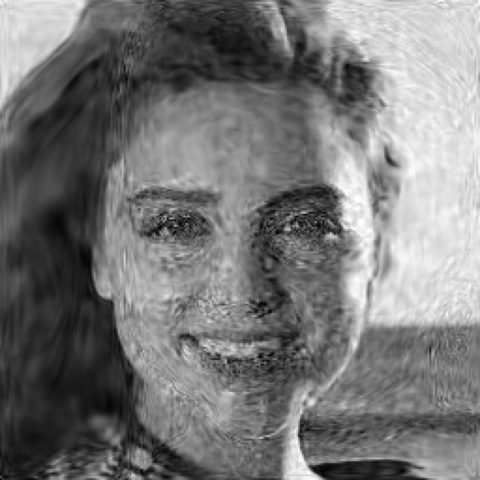} &
    \includegraphics[width=\sza\columnwidth]{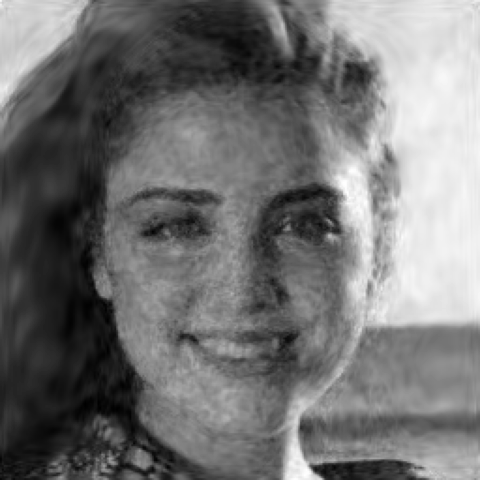} &
    \includegraphics[width=\sza\columnwidth]{figs/gt2.png}
    \\
    \rotatebox{90}{\hspace{12pt}\emph{$0\%$}}
    \hspace{1pt} &
    \includegraphics[width=\sza\columnwidth]{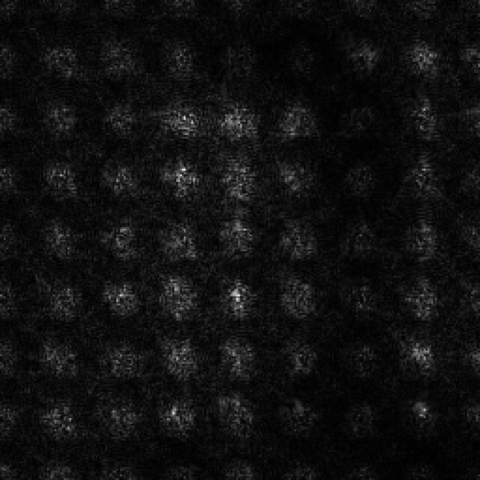} &      
    \includegraphics[width=\sza\columnwidth]{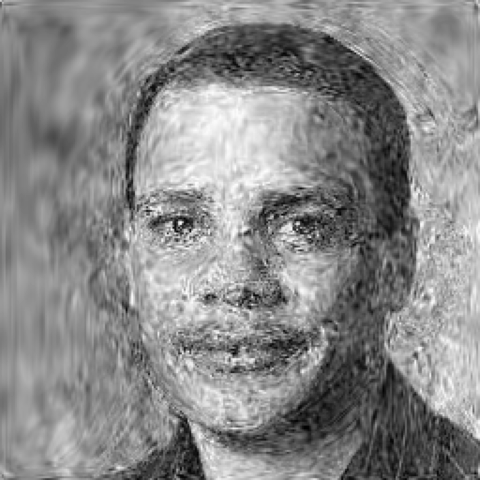} &
    \includegraphics[width=\sza\columnwidth]{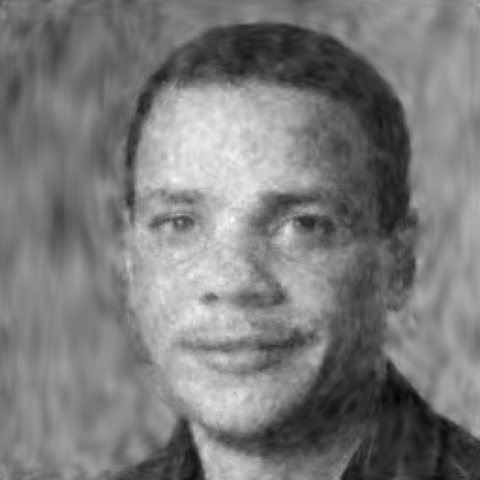} &
    \includegraphics[width=\sza\columnwidth]{figs/gt1.png} &
    \hspace{2pt}
    \includegraphics[width=\sza\columnwidth]{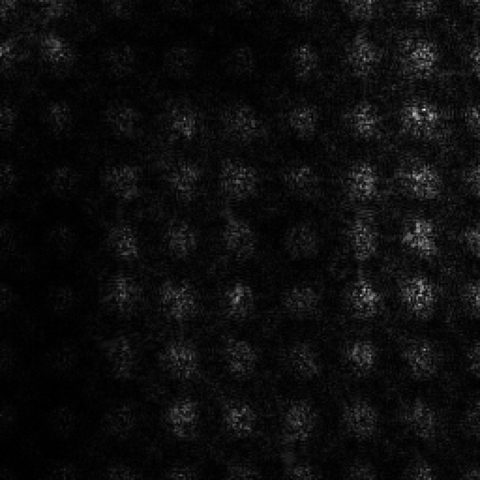} & 
    \includegraphics[width=\sza\columnwidth]{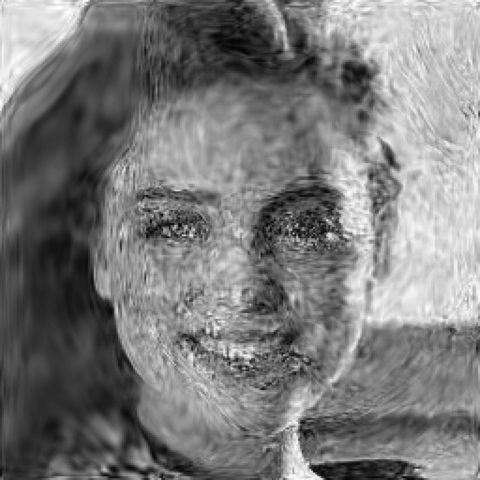} &
    \includegraphics[width=\sza\columnwidth]{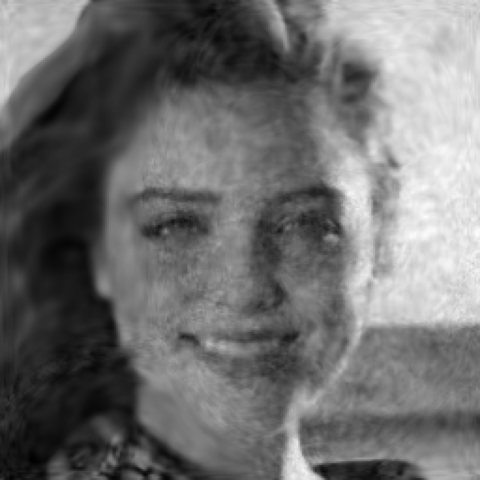} &
    \includegraphics[width=\sza\columnwidth]{figs/gt2.png}
    \\
    \rotatebox{90}{\hspace{12pt}\emph{$-50\%$}}
    \hspace{1pt} &
    \includegraphics[width=\sza\columnwidth]{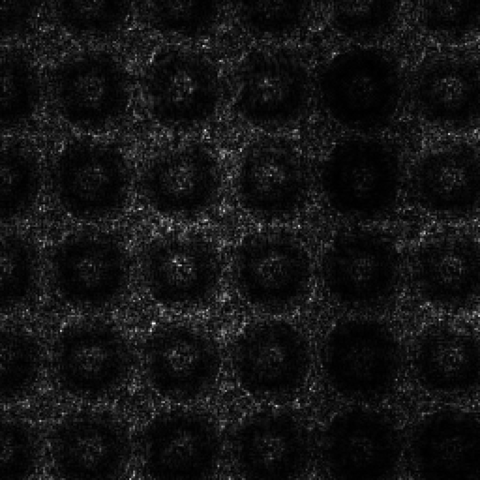} & 
    \includegraphics[width=\sza\columnwidth]{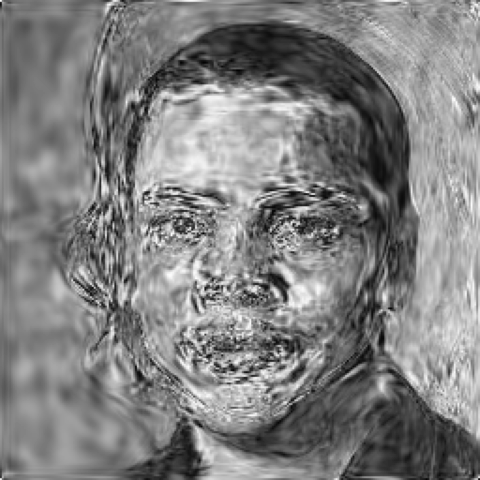} &
    \includegraphics[width=\sza\columnwidth]{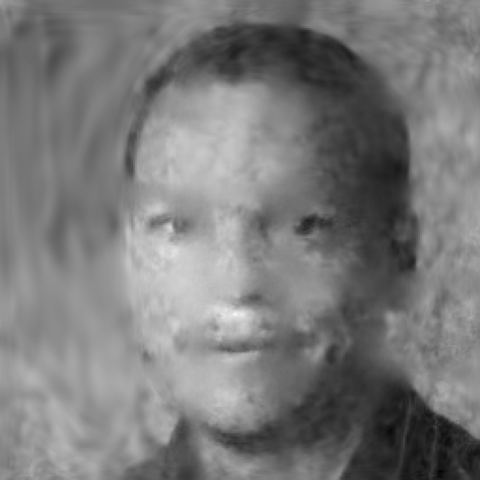} &
    \includegraphics[width=\sza\columnwidth]{figs/gt1.png} &
    \hspace{2pt}
    \includegraphics[width=\sza\columnwidth]{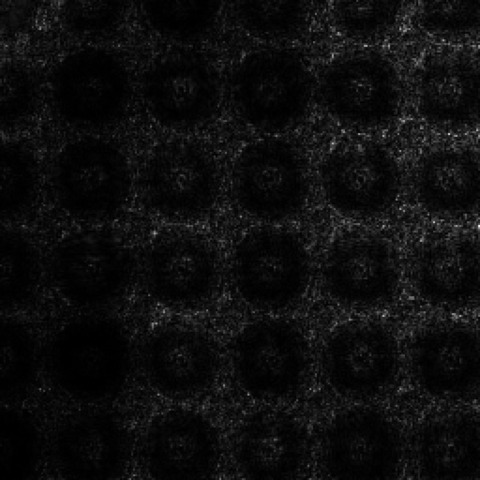} &      
    \includegraphics[width=\sza\columnwidth]{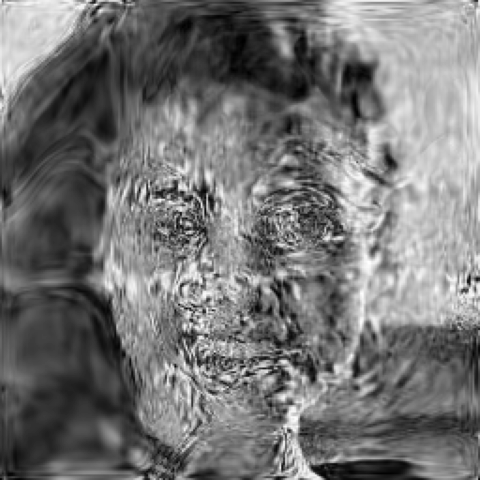} &
    \includegraphics[width=\sza\columnwidth]{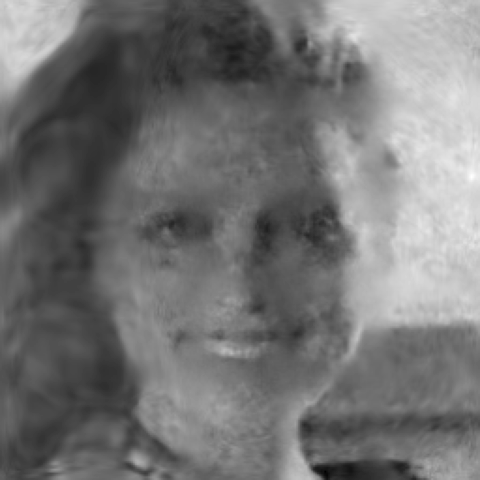} &
    \includegraphics[width=\sza\columnwidth]{figs/gt2.png}

\end{tabular}
\caption{Comparison of the the reconstructions from the proposed method against ePIE and ground truth for noisy case ($\sigma = 2$) for different overlap percentages}
 \label{fig:overlap_ny}
\end{figure*}

In Fig.~\ref{fig:overlap_nf}, the comparison between the ePIE algorithm and the proposed algorithm can be seen for overlap percentages of $75\%$, $50\%$, $25\%$, $0\%$, and $-50\%$ are shown. When we look at the reconstructions with the highest overlap percentage, the reconstructions show similar properties for both algorithms. However, due to the increased computational complexity of the proposed algorithm and the disadvantage of the background for DGPs, the reconstruction from ePIE has slightly better reconstructions. This situation can also be observed in the Table~\ref{tab:tab2}. A similar observations can be made for an overlap percentage of $50\%$. The reconstructions are good for both methods, where the analytical method edges the proposed method with a slight margin due to the imperfect optimization. However, the difference can be eliminated with the correct parameter selection in these cases.

The most easily observable and important result can be seen for the low overlap percentage case, that is $25\%$ overlap. Although the reconstruction quality for the proposed method does not change much, the reconstructions from ePIE are significantly worsens, as can be seen in Fig.~\ref{fig:overlap_nf} and Table~\ref{tab:tab2}. When we move on to no overlap cases, there is no possibility for ePIE to have a reconstruction. On the other hand, the proposed method results in a satisfactory reconstruction with small decreases in the SSIM scores. More importantly, the last three overlap percentages corresponds to approximately $33\%$, $50\%$, and $67\%$ decrease in the scanning time for synchrotron experiments. The decrease in the scanning time results in higher quality scans with less noise in the diffraction patterns, which is crucial in many applications.

A similar situation can be observed for the noisy reconstructions in Fig.~\ref{fig:overlap_ny} and Table~\ref{tab:tab3}. Here noise with relative value 2 ($\sigma = 2$) is used for all reconstructions, and the proposed method is compared with ePIE, where the reconstructions with and without regularization are shown for fairness since ePIE does not have any regularization. 

Reconstructions with ePIE shows observable artifacts in the case of noise. The advantage of solving the problem analytically for ePIE does not apply here since the proposed algorithm is able to generate good reconstructions even without regularization for a low high overlap percentage. Similar to the previous case, with the decreased overlap percentage, ePIE loses the ability to have a reconstruction at all. On the other hand, while the reconstruction quality for the proposed algorithm decreases gradually, the reconstructions without regularization remain good qualitatively. However, the reconstructions show significant artifacts for no overlap cases by the combination of lack of data and the noise, which can also be seen in the SSIM values in Table~\ref{tab:tab3}. 

The addition of regularization increases the reconstruction quality significantly quantitatively, as can be observed from the SSIM values. The qualitative evaluation shows that when there is no overlap, the reconstruction in high noise scenarios might not be acceptable. However, the reconstructions are significantly superior to the ones from ePIE in all cases.

\section{Discussion and Outlook}

In x-ray ptychography, reconstruction quality depends heavily on the overlap percentage between consecutive scanning areas. Due to the limited time allocated for a sample to be scanned, the overlap amount cannot be kept high in all experiments. Similarly, with the increased overlap, the dwelling time for each scanning position is shortened, leading to noisier diffraction patterns recorded at the detector. The trade-off between these aspects leads to imperfect reconstructions in x-ray ptychography, which either do not have enough redundant data for good reconstructions or show artifacts related to noise. Existing methods in the literature approach the problem with different methods. While some try to increase the reconstruction quality by modifying the algorithm, others try increasing the output resolution via post-processing. However, these are not always good enough for some experiments.

Usage of deep generative priors (DGPs) gives an important advantage for achieving good reconstructions. Since the starting point for any ill-posed problem is vital for a good solution, having a good initialization is crucial for x-ray ptychography also. Usages of DGPs do not only enable us to start with a better initial reconstruction, but it also provides a network that is trained to give reconstructions in a constrained domain, which adds additional priors to the problem. Thus, the overlapping requirement is significantly reduced even just by the effect of DGPs.

\setlength\tabcolsep{3.2pt}
\begin{table}[!b]
\footnotesize
\caption{Structural Similarity Index (SSIM) comparison for ePIE, proposed method without regularization (Prop), and proposed method with regularization (Prop w/ R) methods and different overlap percentages using noisy data ($\sigma = 2$)}
\centering
\begin{tabular}{@{}c|c|c|c||c|c|c|@{}}
\multicolumn{1}{|c|}{\textbf{SSIM}} &
\multicolumn{3}{c||}{\textbf{Image 1}} & \multicolumn{3}{c|}{\textbf{Image 2}} \\
% \cmidrule(l){2-9} 
\midrule
\multicolumn{1}{|c|}{Method} & {ePIE} & {Prop} & {Prop w/ R}  & {ePIE} & {Prop} & {Prop w/ R} \\ \midrule
\multicolumn{1}{|c|}{Overlap = $75\%$} & 0.50 & 0.93 & \textbf{0.95} & 0.51 & 0.87 & \textbf{0.93} \\
\multicolumn{1}{|c|}{Overlap = $50\%$} & 0.54 & 0.86 & \textbf{0.90} & 0.54 & 0.84 & \textbf{0.90} \\
\multicolumn{1}{|c|}{Overlap = $25\%$} & 0.26 & 0.73 & \textbf{0.85} & 0.22 & 0.75 & \textbf{0.87} \\
\multicolumn{1}{|c|}{Overlap = $0\%$} & 0.17 & 0.60 & \textbf{0.84} & 0.15 & 0.65 & \textbf{0.84} \\
\multicolumn{1}{|c|}{Overlap = $-50\%$} & 0.16 & 0.52 & \textbf{0.80} & 0.15 & 0.52 & \textbf{0.75} \\
\hline
\end{tabular}
\label{tab:tab3}
\end{table}

In addition to GDPs, the reconstructions can be improved significantly with the usage of deep image priors (DIPs). DIPs allow a solution to be optimized using the structure of a network. Optimizing the weights and filters of a network with a single set of data specializes the network for a specific solution; in other words, the network over-fits the data. However, during this process, due to the abundance of network parameters, the initial points of the network weights and the network structure plays an important role in the reconstructions. Due to this advantage, the combination of DGPs with DIPs generates good reconstructions even though there is not enough data for an exact solution. The initial priors learned by DGPs, and the structural priors learned by DIPs constrain solutions to a smaller set of solutions with significantly lower error. The advantage can be best seen in Fig.~\ref{fig:overlap_nf} and Table~\ref{tab:tab2}.

The combination of the DGPs and DIPs can be improved further based on the properties of the objects to be reconstructed. In case of a noisy measurements, other priors can be utilized as regularizers, such as total variation (TV) regularization and discriminator loss (DL) regularization. TV regularizer forces the solution to be piece-wise smooth, which eliminates point-like artifacts caused by noisy data in the reconstructions as it is used in similar approaches. Moreover, the generator network is trained together with a discriminator network which specializes in classifying if an object belongs to a certain set of samples or not. This information can be used to punish reconstructions diverging too much from the desired range of outputs. Essentially, the regularizer pushes the solution towards a general understanding of a face image in our examples, which is an averaged out face concept. If the output is too noisy and has a lot of artifacts, it will be far from this range and will be punished by the regularizer. The effect of TV and DL regularizers on top of DIPs and DGPs can be best observed in the noisy reconstructions in Fig.~\ref{fig:overlap_ny} and Table~\ref{tab:tab3}

Despite the improved quality of reconstructions, the most important drawback of the proposed method is the reconstruction time. The method is a self-optimizing algorithm, and the network is needed to be optimized for each reconstruction. That makes the proposed method considerably slower compared to the iterative reconstruction methods. In our simulations, the training is done on an Nvidia GeForce Titan X graphics card, where reconstruction with $50\%$ overlap can take 23 minutes, and reconstruction with $25\%$ overlap takes 15 minutes. The reconstruction time may increase or decrease based on the parameters chosen for optimization. Although that is an important drawback, the reconstruction time can be improved significantly by different methods. The largest improvement can be made by sharing the computational burden among multiple GPUs. Since the ptychographic process has multiple scanning points, scanning points can be scattered randomly among GPUs for improved speed. Also, all computations can be split among GPUs without high-level distribution, which can help overcome this bottleneck \cite{Hidayetolu2019, Majchrowicz2020}. 

An improvement in the reconstructions for low overlap scenarios can be made by a combination of an analytical method (such as ePIE) and the proposed algorithm for a simpler but more accurate reconstruction. In this method, reconstructions from insufficient overlap data can be made using the proposed algorithm, and new data can be acquired from the reconstruction with a higher overlap percentage using the forward model. If the low overlap data is combined with this generated high overlap data, an analytical method can be used for a more quality reconstruction.
\section{Conclusion}

In this paper, we have proposed an x-ray ptychography reconstruction method utilizing deep image priors, deep generative priors, and other data-based priors. We demonstrated that using deep image priors to optimize the solution for the highly ill-posed x-ray ptychography problem can generate improved reconstruction results. We also showed that using a pre-trained generator on a dataset that is similar to the target reconstruction can greatly improve the initial state for the deep image priors, drive the solution towards the desired domain, and provide an implicit prior for a quality reconstruction in a shorter time compared to a similar deep image priors method. Moreover, we have shown that combining the method total variation prior or a prior coming from the discriminator of the used generator, and using these priors as regularizers would lower the artifacts caused by noisy data. Finally, we demonstrated the ability of the proposed method, particularly in low-overlap data, to have significantly improved reconstruction quality compared to analytical methods.
\section*{Acknowledgments}
This research used resources of the Advanced Photon Source, a U.S. Department of Energy (DOE) Office of Science User Facility at Argonne National Laboratory, and is based on research supported by the U.S. DOE Office of Science-Basic Energy Sciences under contract DE-AC02-06CH11357.
% \nocite{*}
\bibliographystyle{IEEEtran}
\bibliography{8_Bibliography/bibliography.bib}
\vspace{-30pt}
\begin{IEEEbiography}[{\includegraphics[width=1in,height=1.25in,clip,keepaspectratio]{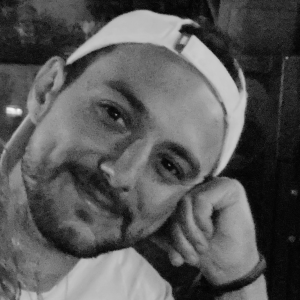}}]{Semih Barutcu}
received his B.Sc. degree in Electrical and Electronics Engineering from Bogazici University, Istanbul, Turkey, and his M.Sc degree in Electrical Engineering and Computer Science from Northwestern University, Evanston, IL, USA, in 2017 and 2018, respectively. He is currently working toward a Ph.D. degree at Northwestern University, Evanston, IL, USA. His research interests include computational imaging, deep learning, and inverse problems in computer vision.
\end{IEEEbiography}

\vspace{-30pt}
\begin{IEEEbiography}[{\includegraphics[width=1in,height=1.25in,clip,keepaspectratio]{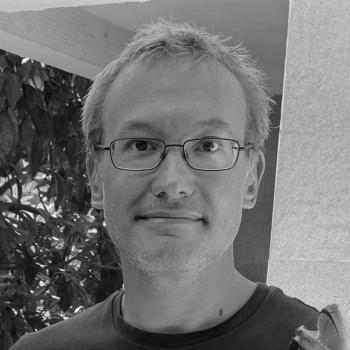}}]{Do\u ga G\" ursoy}
is a computational scientist at the X-ray Science Division of Advanced Photon Source. He is also an adjunct associate professor in the Electrical Engineering and Computer Science Department at Northwestern University, and a fellow of the Northwestern Argonne Institute for Science and Engineering. Dr. Gursoy’s research focuses on computational imaging and inverse problems in imaging sciences. He is an associate editor for the IEEE Transactions on Computational Imaging, and a member of IEEE Signal Processing Society’s Technical Committee on Computational Imaging.
\end{IEEEbiography}

\vspace{-30pt}
\begin{IEEEbiography}[{\includegraphics[width=1in,height=1.25in,clip,keepaspectratio]{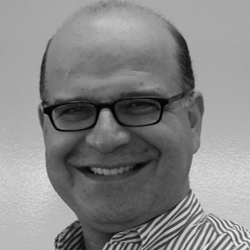}}]{Aggelos K. Katsaggelos}
is Professor and Joseph Cummings Chair in the ECE Department at Northwestern University. He also runs the Image and Video Processing Laboratory (IVPL), whose objective is to generate cutting-edge research results in the fields of multimedia signal processing, multimedia communications, and computer vision. IVPL works in a variety of problems (e.g., recovery, compression, segmentation, and speech and speaker recognition) and applications areas (e.g., medical, multi-spectral, and astronomical image processing). Dr. Katsaggelos is a Fellow of the IEEE (1998) and SPIE (2009), the co-inventor of seventeen international patents, the recipient of the IEEE Third Millennium Medal (2000), the IEEE Signal Processing Society Meritorious Service Award (2001), the IEEE Signal Processing Society Technical Achievement Award (2010), and co-author of several award-winning papers.
\end{IEEEbiography}

% \vfill

\end{document}